\documentclass[lettersize,journal]{IEEEtran}

\usepackage{amsmath,amsfonts,amssymb}
\usepackage{algorithmic}
\usepackage{algorithm}
\usepackage{array}
\usepackage{multirow}
\usepackage{makecell}
\usepackage{rotating}
\usepackage{graphicx}

\usepackage{subcaption}

\usepackage{textcomp}
\usepackage{stfloats}
\usepackage{url}
\usepackage{verbatim}
\usepackage{cite}
\usepackage{xcolor}
\usepackage{xspace}
\usepackage{microtype}

\usepackage[T1]{fontenc}
\usepackage[utf8]{inputenc}

\usepackage{xspace}

\newcommand{\eg}{\emph{e.g.}\xspace}

\begin{document}

\title{Detecting Hallucination in LLMs: Tracing the Topological Signatures of Impaired Context Sharing}

\author{
Amir Jalilifard,
Anderson Rocha,
Eric Wong,
and Marcos Medeiros Raimundo%
\thanks{Amir Jalilifard is with the Institute of Computing,
Universidade Estadual de Campinas (UNICAMP), Campinas, SP, Brazil.
E-mail: jalilifard@ic.unicamp.br.}%
\thanks{Anderson Rocha is with the Institute of Computing,
Universidade Estadual de Campinas (UNICAMP), Campinas, SP, Brazil.
E-mail: anderson.rocha@unicamp.br.}%
\thanks{Eric Wong is with the Department of Computer and Information Science,
University of Pennsylvania, Philadelphia, PA, USA.
E-mail: exwong@cis.upenn.edu.}%
\thanks{Marcos Medeiros Raimundo is with the Institute of Computing,
Universidade Estadual de Campinas (UNICAMP), Campinas, SP, Brazil.
E-mail: mrai@unicamp.br.}%
}


\markboth{Journal of \LaTeX\ Class Files,~Vol.~14, No.~8, August~2021}%
{Shell \MakeLowercase{\textit{et al.}}: A Sample Article Using IEEEtran.cls for IEEE Journals}


\maketitle

\begin{abstract}
In this work, we examine the topology of information flow patterns within attention graphs to effectively distinguish hallucinated from non-hallucinated responses. We analyze the Forman–Ricci curvature to identify structural patterns indicating information bottlenecks in attention graphs. We then introduce a method that captures both semi-local and global information-flow characteristics of attention heads associated with hallucinated responses. We evaluate our approach extensively across several LLMs and established benchmarks. Empirical results demonstrate that our proposed single-pass approach provides consistent improvements over existing attention-based and multi-response baselines across two hallucination-detection benchmarks, while achieving competitive performance across diverse LLM architectures. Further analysis reveals that impaired context sharing among tokens during causal generation is strongly associated with hallucination occurrences in LLMs. In particular, hallucinated responses are consistently characterized by an over-reliance on self-attention, diffused context retrieval from earlier tokens, or information over-squashing, especially in the final transformer layer.
\end{abstract}

\section{Introduction}

Amid the advances and extensive use of LLMs in various domains, new challenges regarding the factuality and correctness of LLM responses have emerged \cite{tao2024trust}. While these models are increasingly integrated into broader autonomous workflows and downstream applications \cite{shinn2023reflexion, yao2023react}, ensuring the factual precision of individual generated outputs remains a fundamental bottleneck for their reliability. A single unfaithful or factually incorrect response can compromise user trust and limit a model's practical utility. Misleading outputs generally fall into two categories: factual errors and faithfulness errors \cite{berglund2023reversal}. These errors may arise from reasoning failures \cite{berglund2023reversal}, defective inference \cite{chen2022towards}, architectural limitations \cite{huang2025survey}, training data biases, or inconsistencies in the provided context \cite{maynez2020faithfulness}. Consequently, isolating and detecting factual deviations within single-turn generations is critical for secure deployment.

Recent research indicates that hallucination, defined as the generation of factually incorrect or unsubstantiated responses, correlates with confusion and instability in the model during response generation. As a result, verifying the self-consistency of LLMs has become an important strategy for detecting hallucination \cite{farquhar2024detecting, wang2022self}. Although this methodology has demonstrated promising results, it possesses two significant limitations. Firstly, it is computationally expensive because it requires generating multiple responses to a single prompt \cite{chen2024inside}. Secondly, this approach offers limited insight into the underlying causes and mechanisms behind these uncertainties.

Building upon literature that interprets attention weights as a weighted adjacency matrix \cite{barbero2024transformers}, we evaluate topological properties of the attention graph that can impede information flow. Consequently, we propose a set of topological features derived from attention weights that effectively distinguish hallucinated from factual responses. The main contributions of our work are as follows:

\begin{itemize}
    \item We demonstrate that LLM hallucinations are characterized by information bottlenecks in the attention mechanism, which we identify using Ricci curvature on the attention graph.
    \item We propose a novel set of topological features from attention maps that enable a simple linear probe to consistently outperform strong attention-based and multi-response baselines.
    \item We provide a comparative analysis across three LLM families, showing that the specific topological signatures of hallucination vary with model architecture, even though the core phenomenon of information bottlenecks is consistent.
\end{itemize}

\section{Background} \label{sec:background}

In this section, we establish the conceptual foundations for our topological analysis of Large Language Models (LLMs). We first define the formal mechanism of attention as a context-sharing process. Subsequently, we discuss how structural bottlenecks within attention graphs lead to information over-squashing. Finally, we introduce the Forman-Ricci curvature as a mathematical framework for quantifying and detecting these bottlenecks using graph topology.

\subsection{Attention Formalism}
\label{subsec:attention_formalism}

In attention-based language models, the entire attention generation can be seen as a context-sharing process. Let \begin{math} \left\{ T_1, T_2, \dots, T_{n}  \right\} \end{math} be a sequence of tokens, \begin{math} A = (a_{1}, a_{2}, \dots a_{n} ) \end{math} denote attention score vector, and \begin{math} V = \left\{v_{1}, v_{2}, \dots v_{n} \right\} \end{math} the matrix of value vectors carries the token contextual information. The context vector for the nth token is therefore
$\text{C}_n = \sum_{j=1}^{n} a_j \cdot v_j$. This context vector determines how much information from previous tokens contributes to the representation of token \textit{n}. Although attention weights reference previous tokens, the overall information flows forward through the sequence. 

\subsection{Information Bottlenecks and Over-squashing}
\label{subsec:bottlenecks}

\begin{figure}[h]
\centering
\includegraphics[width=0.48\textwidth]{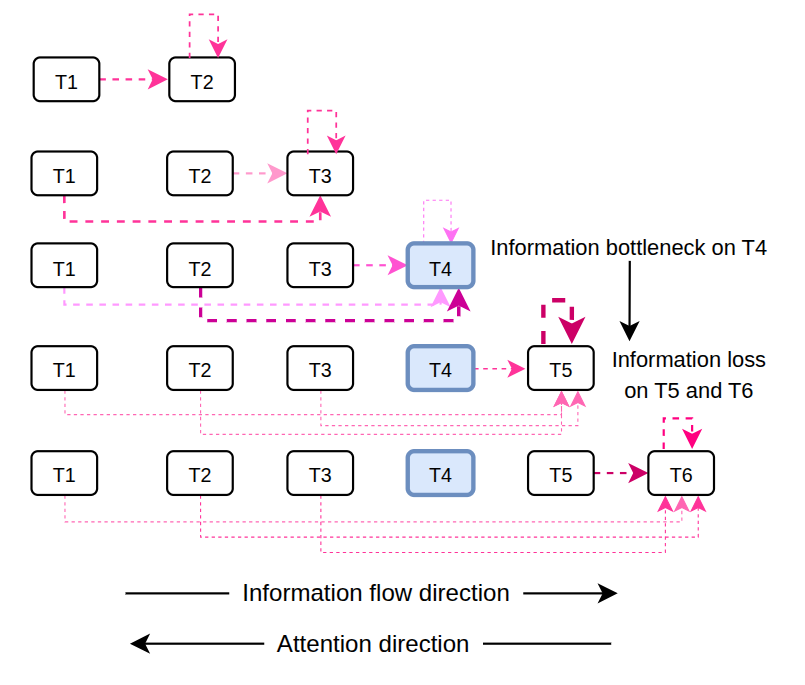}
\caption{Information over-squashing occurs when information from many tokens is funneled through only a single token that bears the burden of representing all that context. However, this central token and its predecessors receive insufficient attention from subsequent tokens, preventing effective context sharing. In this example, all the information passes through token \textit{T4}, but later, due to high self-attention on \textit{T5}, a significant amount of information from earlier tokens is lost. The line thickness and its color show the amount of information flow from previous tokens to the subsequent ones.}
\label{fig:bottleneck}
\end{figure}

A bottleneck happens when a significant part of the information from previous tokens passes through a single token. Such a narrow information flow leads to two main problems. First, concentrating the context of multiple tokens into a single token leads to excessive information compression, which can cause loss of contextual details. This effect is particularly pronounced in long sequences containing dozens and hundreds of tokens. Second, as illustrated in Figure \ref{fig:bottleneck}, even if a single token carries comprehensive context from previous tokens, this information can subsequently be lost if attention becomes overly focused on a small subset of tokens further along the sequence, or if a token predominantly attends to itself.

An attention matrix can be seen as a weighted graph, with tokens as vertices and attention weights as edges. Information flows between tokens and attention layers via a message-passing mechanism. Distortion of the information flow due to bottlenecks limits the efficiency of message-passing. This phenomenon, called information over-squashing, is frequently associated with information bottlenecks in graphs and is measured by Forman-Ricci curvatures \cite{forman2003bochner, alon2020bottleneck, topping2021understanding}.

\subsection{Forman-Ricci Curvature}
\label{subsec:forman_ricci}

The presence of bottlenecks in attention graphs strongly co-occurs with context loss, disrupting the coherent semantic construction expected in transformer models. The Forman-Ricci Curvature (FRC)~\cite{forman2003bochner, sreejith2016forman} provides a measure of information flow passing through an edge \( e = (v, u)\). It consists of two components: the positive part, which measures the direct connections, and the penalty part, which measures the triangular alternative connection between two nodes. Therefore, \textbf{positive} expected value of FRCs of a graph indicates alternative information flux rather than passing through edge \( e \), and \textbf{negative} values indicate information bottleneck, i.e., the information route between \( v \) and \( w \)  is focused on the edge \( e \). Since attention graphs are directed weighted graphs, we denote the attention weight from token \(u\) to token \(v\) by \(w_{uv}\).

Let \( G = (V, E) \) be the attention graph, with vertices \( v \) and \( w \) connected by an edge \( e \). FRC is defined as:

\begin{equation}
\begin{split}
\operatorname{Forman-Ricci}(v,u) 
=
w_{uv} \Biggl(
\frac{w_{v}}{w_{uv}}
+
\frac{w_{u}}{w_{uv}}
\; - \;\\
\sum_{\substack{y \sim N(v) \setminus u \\ z \sim N(u) \setminus v}}
\left[
\frac{w_{v}}{\sqrt{w_{uv} \, w_{vy}}}
+
\frac{w_{u}}{\sqrt{w_{uv} \, w_{uz}}}
\right]
\Biggr)
\label{eq:frc}
\end{split}
\end{equation}


where \(w_{uv}\) denotes the attention weight associated with the directed edge \((u,v)\), and

\[
w_u = \sum_{y \sim N(u)} w_{uy}
\]

denotes the weighted out-degree of node \(u\). The neighborhoods \(N(u)\) and \(N(v)\) denote the sets of vertices connected to \(u\) and \(v\), respectively.

\section{Methodology}

\subsection{Topological Indexes of Information Flow}

To quantify the efficiency of context sharing within attention graphs, we define three specific topological and information-theoretic indices.

\subsubsection{Mutual Outgoingness}
In its simplified form, when the penalty term is omitted, the Forman--Ricci curvature reduces to the sum of the weighted degrees of the two connected nodes. Let

\[
d(u) = \sum_{v \sim N(u)} w_{uv}
\]

denote the weighted out-degree of token \(u\). In attention graphs, \(d(u)\) measures how broadly a token distributes attention across neighboring tokens.

We therefore define the mutual outgoingness score of an edge \((u,v)\) as

\[
\operatorname{Outgoingness}(u,v)=d(u)+d(v).
\]

Intuitively, low outgoingness values indicate highly self-focused attention patterns, whereas high outgoingness values indicate broader context sharing across tokens. The score can be computed efficiently in \(\mathcal{O}(E)\).

\subsubsection{Forman-Ricci Penalty}
In contrast, the full Forman-Ricci curvature includes a penalty term that reflects edge redundancy and alternative paths, and is directly tied to local information flow. The penalty part of FRC of a weighted graph from (\ref{eq:frc}) can be written in terms of probabilities:

\begin{equation}
\begin{split}
P(u,v) = \sum_{y \sim N(v) \setminus u}  \frac{1}{\sqrt{p_v(u) p_v(y)}} \\ + \sum_{z \sim N(u) \setminus v}\frac{1}{\sqrt{p_u(v) p_u(z)}}
\label{eq:pen-u-v}
\end{split}
\end{equation}
where $p_u(v) = \frac{w_{uv}}{w_u}$, $p_v(u) = \frac{w_{uv}}{w_v}$ and so on.

The function $\frac{1}{\sqrt{p_u(v) p_u(z)}}$ is convex over the probability simplex with respect to $p_u$. Because convexity is preserved under summation, the penalty term is minimized when the attention distribution is uniform \cite{bauschke2022minimal}.

\subsubsection{Joint Entropy as an FRC Penalty Proxy}

From the chain rule of entropy, we can write:
\begin{equation}
H(U, V) = H(U) + H(V \mid U)
\label{eq:chain-rule}
\end{equation}
where

\begin{equation}
H(V \mid U) = -\sum_{u \in V} p(u) \sum_{v \sim N(u)} p(v \mid u) \log p(v \mid u)
\label{eq:conditional-entropy}
\end{equation}

assuming that $p_u(v) = \frac{w_{uv}}{w_u}$, and $p(u|v) = \frac{p(u,v)}{p(v)}$ and $p(u,v) = \frac{w_{uv}}{W_{\text{total}}}$, then

\begin{equation}
p(v \mid u) = \frac{p(u, v)}{p(u)} = \frac{\frac{w_{uv}}{W_{\text{total}}}}{\frac{w_u}{W_{\text{total}}}} = \frac{w_{uv}}{w_u} = p_u(v)
\end{equation}

Therefore,

\begin{equation}
H(V \mid U) = -\sum_{u \in V} p(u) \sum_{v \sim N(u)} p_u(v) \log p_u(v)
\label{eq:conditional-entropy-puv}
\end{equation}

From equations (\ref{eq:pen-u-v}), (\ref{eq:chain-rule}), and (\ref{eq:conditional-entropy-puv}), a uniform $P_u(v)$ maximizes the joint entropy while it minimizes the total penalty, indicating a possible reverse correlation between $H(u, v)$ and total penalty in Forman-Ricci. Both the entropy formulation and the Forman--Ricci penalty term are optimized under more uniformly distributed attention weights. This motivates the use of attention entropy as a computationally efficient proxy for curvature-related information dispersion and bottleneck behavior.


\subsection{Extracting Features for Hallucination Detection}

The exact computation of the full Forman-Ricci penalty incurs a worst-case complexity of $\mathcal{O}(V \cdot \bar{d}^2)$, where $\bar{d}$ is the average node degree, making it costly for long sequences or densely connected graphs. Given that attention graphs often span hundreds or thousands of tokens, using raw edge-level curvatures can be noisy and overly granular. More importantly, such local curvatures make the final set of features for hallucination detection sentence-length dependent.

To address this and create a set of meaningful, fixed-size features as the probe's input, we summarize the distribution of mutual outgoingness scores in each attention head by computing the 10th, 50th, and 90th percentiles, capturing the distribution of local connectivity patterns within each attention head. Additionally, motivated by the inverse relationship between curvature penalties and joint Shannon entropy $H(U, V)$, and its linear time complexity of $\mathcal{O}(V)$, we include the joint entropy of each attention head as a cheap-to-calculate feature to characterize global information flow and identify systemic information bottlenecks associated with hallucinated generations.

\section{Experiments}
In this section, we describe the experimental setup, report the analytical results, and present the outcomes of hallucination detection and discuss the effectiveness of our approach.

\subsection{Experimental Setup}
We carried out all the experiments using three small-to-medium-sized models — Mistral 7B Instruct, LLaMA 3.1 8B Instruct, and Phi-4—and one larger model, Qwen3-32B. In particular, we included Qwen3-32B as a larger reasoning-oriented model to examine whether reasoning-specific mechanisms (e.g., explicit intermediate reasoning steps) influence characteristics such as outgoingness and attention entropy. For each model–benchmark pair, we generated responses using two sampling temperatures (0.1 and 1.0). We assessed the performance of our approach using two benchmarks commonly employed in hallucination studies: TruthfulQA \cite{lin2021truthfulqa} and NQ-Open \cite{lee2019latent}. The TruthfulQA comprises 817 questions and ground-truth answers specifically designed to elicit imitative falsehoods and mimic human misconceptions. It contains 38 categories, including health, law, conspiracies, and fiction. The NQ-Open is a dataset of realistic questions collected from Google Search and consists of 3,610 questions also accompanied by ground-truth answers.

Both LLaMA 3.1 8B and Mistral 7B Instruct contain 32 layers, each with 32 attention heads, whereas Phi-4 has 40 layers with 40 heads per layer. Qwen3-32B with 64 layers and 64 heads per layer, has the largest architectural depth and width among the evaluated models. The experimental setup, including system prompt templates, attention implementation (Eager attention), hallucination probe,  and dimensionality reduction settings, follows the exact configuration in \cite{binkowski2025hallucination}, with the exception of label generation, where we used GPT-4 instead of GPT-4 mini due to its superior ability to capture nuanced linguistic patterns and provide significantly more accurate labels.

The use of an LLM-as-a-judge was motivated by the generation of tens of thousands of responses with complex linguistic variations, making exhaustive human evaluation impractical. To assess the reliability of the judge, we randomly sampled 100 prompts and manually compared the generated labels against the reference answers. The evaluation showed that GPT-4 achieved accuracies of 92.0\%±3.34\%, 91.0\%±5.61\%, 88.0\%±6.4\%, and 87.0\%±6.6\% (95\% confidence intervals) when evaluating responses generated by LLaMA 3.1 8B Instruct, Phi-4, Mistral 7B Instruct, and Qwen3-32B, respectively. Since generated responses can contain complex and nuanced information, determining whether a response should be classified as a hallucination may involve subjective judgment. For instance, a model may initially provide a correct answer but later introduce an unsupported or fabricated detail during the continuation of its response. In such cases, due to our strict evaluation criteria, we consider the entire response to be hallucinated even if it is a minor portion of the overall response, whereas an LLM-as-judge may assign a correct label.

In all experiments, the Mann–Whitney U test was used to assess the statistical significance of differences in per-head features. We show the analytics results only for the 10th percentile since the 50th and 90th percentiles follow the same patterns.

\subsection{Forman-Ricci curvature patterns}
We first evaluate whether there are distinct information-flow patterns in Forman–Ricci curvature that differentiate hallucinated answers from the rest. To this end, the LLM responses for each of the NQ-Open questions are partitioned into hallucinated and non-hallucinated subsets. For each subset, we then calculate the mean curvature of each attention head in the last five transformer layers, as the last few layers are known to capture higher-level semantic representations \cite{zhao2024explainability}.

\begin{figure}[h]
  \centering
    \includegraphics[width=0.48\textwidth]{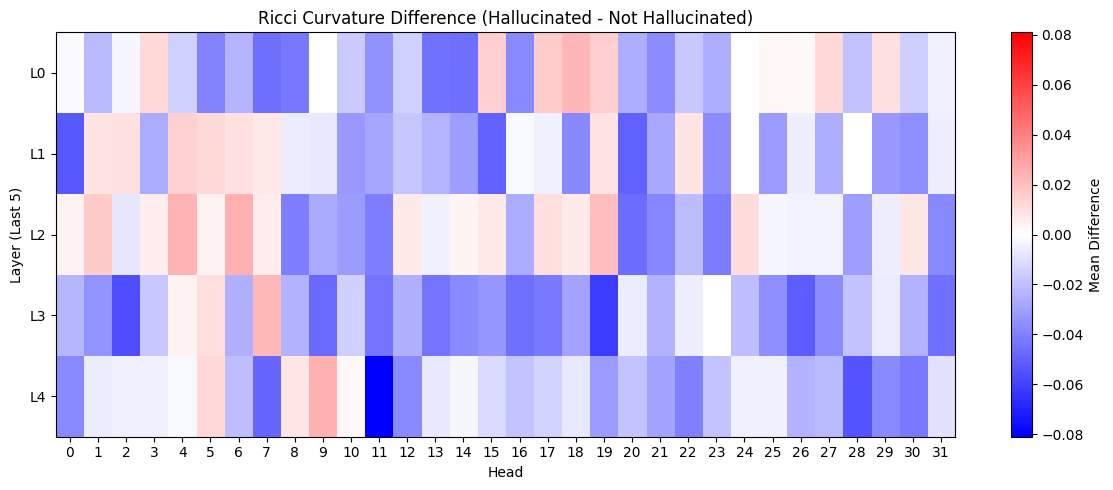}
    \label{fig:ricci-LLaMA}

  \caption{The difference between the mean values of Forman-Ricci curvatures per head for hallucinated and not hallucinated responses for LlaMA. For the majority of the attention heads, hallucinated responses have lower curvature values, indicating a less fluid information flow.}
  \label{fig:ricci-differences-Lamma}
\end{figure}

Figure \ref{fig:ricci-differences-Lamma} presents the average per-head differences in Forman–Ricci curvature between hallucinated and non-hallucinated responses for LLaMA. Negative values (shown in blue) indicate that hallucinated responses tend to exhibit lower curvature values, suggesting more obstructed information flow due to the higher penalty. Across all models, the last layer exhibits less fluid information flow in hallucinated responses. This effect is particularly pronounced in LLaMA and Mistral, where the distinction is more consistent across layers and heads. For Phi and Qwen, however, only the majority of heads in the last layer show attenuated information flow for hallucinated responses. The complete results and statistical significance values are presented in the Appendix, Figures \ref{fig:ricci-differences} and \ref{fig:p-values-ricci-differences}.

\subsection{Outgoingness, entropy and self-focus patterns}

To investigate how outgoingness, distribution of attention across preceding tokens, and self-attention correlate with hallucinated responses, we compared per-head averages of the 10th, 50th, and 90th percentiles of outgoingness scores, attention entropy, and self-attention values across all layers.

As illustrated in Figure \ref{fig:LLaMA-pvals} (also in Appendix, Figure \ref{fig:features-pvalues-part1}), LLaMA 3.1 8B Instruct shows the strongest separation of outgoingness and attention entropy among the other models. In turn, Phi-4 shows a statistically significant difference between the means of hallucinated and non-hallucinated responses for a subset of attention heads, while exhibiting almost no separation of attention entropies for most heads. Mistral 7B Instruct also exhibits meaningful differences among the majority of heads for both outgoingness and attention entropy, though the effect is less pronounced than in LLaMA 3.1 8B Instruct. Finally, Qwen3-32B has the fewest attention heads and shows statistically significant differences in outgoingness and attention entropy.

\begin{figure}[h]
  \centering
  \begin{subfigure}[b]{0.23\textwidth}
    \includegraphics[width=\linewidth]{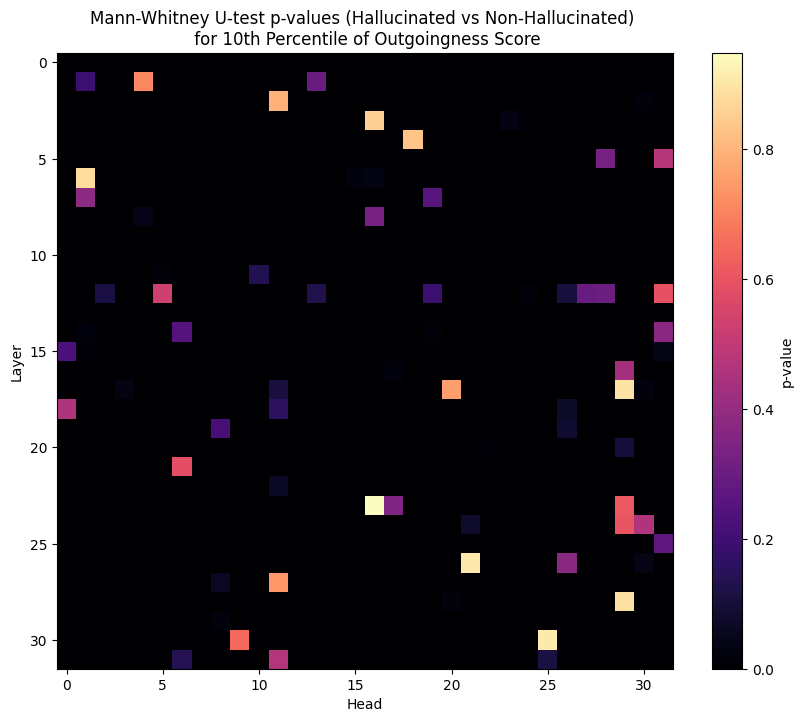}
    \caption{Outgoingness - LLaMA 3.1 8B Instruct}
    \label{fig:10th-percentile-LLaMA-single}
  \end{subfigure}
  \begin{subfigure}[b]{0.23\textwidth}
    \includegraphics[width=\linewidth]{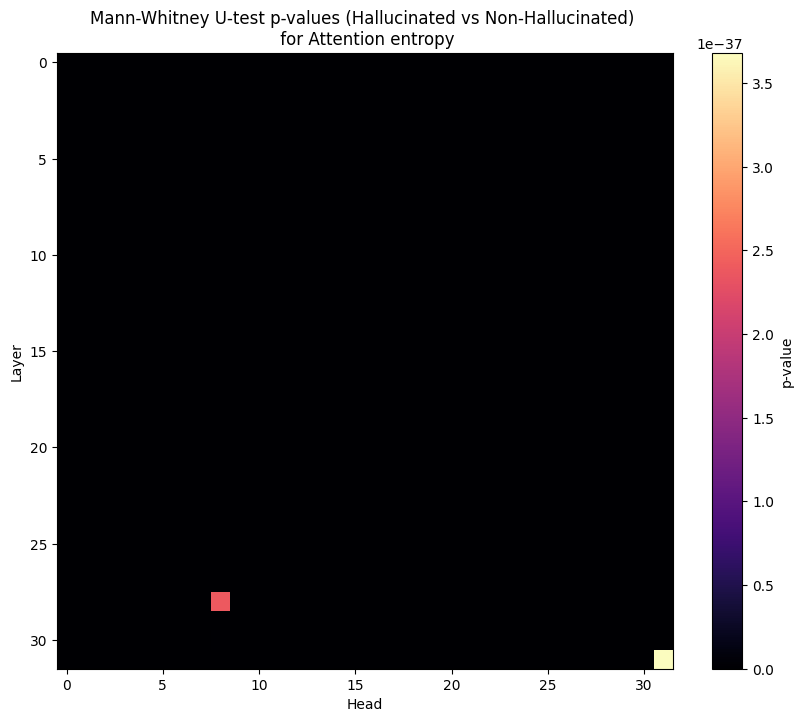}
    \caption{Attention entropy - LLaMA 3.1 8B Instruct}
    \label{fig:attention-entropy-LLaMA-single}
  \end{subfigure}
  \caption{P-values for the per-head comparisons of the 10th percentile outgoingness scores and attention entropy across all layers for LLaMA 3.1 8B Instruct.}
  \label{fig:LLaMA-pvals}
\end{figure}

\begin{figure}[h]
  \centering
  \begin{subfigure}[b]{0.48\textwidth}
    \includegraphics[width=\linewidth]{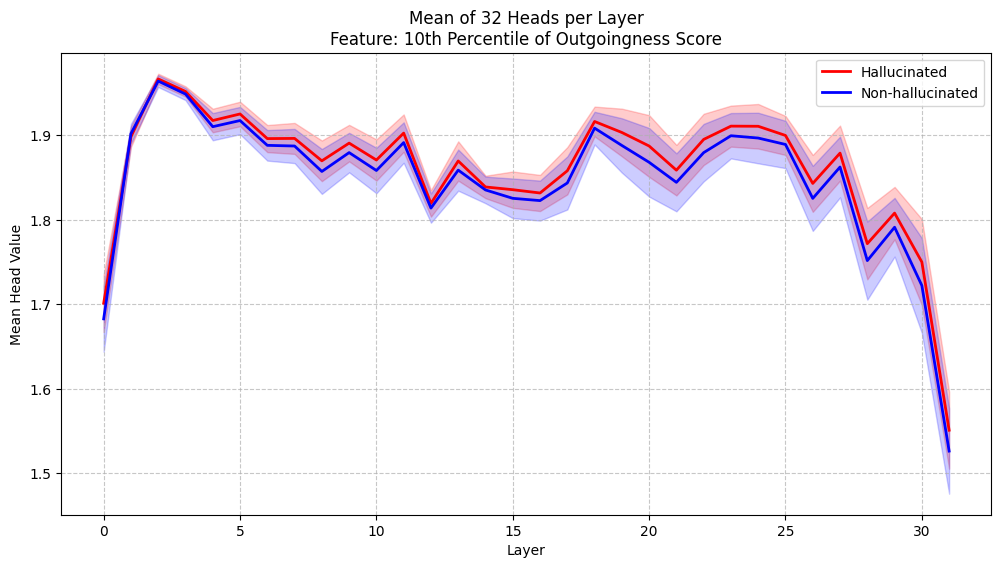}
    \caption{Outgoingness - LLaMA 3.1 8B Instruct}
    \label{fig:LLaMA-10th-percentile-layer-wise-single}
  \end{subfigure}
  \begin{subfigure}[b]{0.48\textwidth}
    \includegraphics[width=\linewidth]{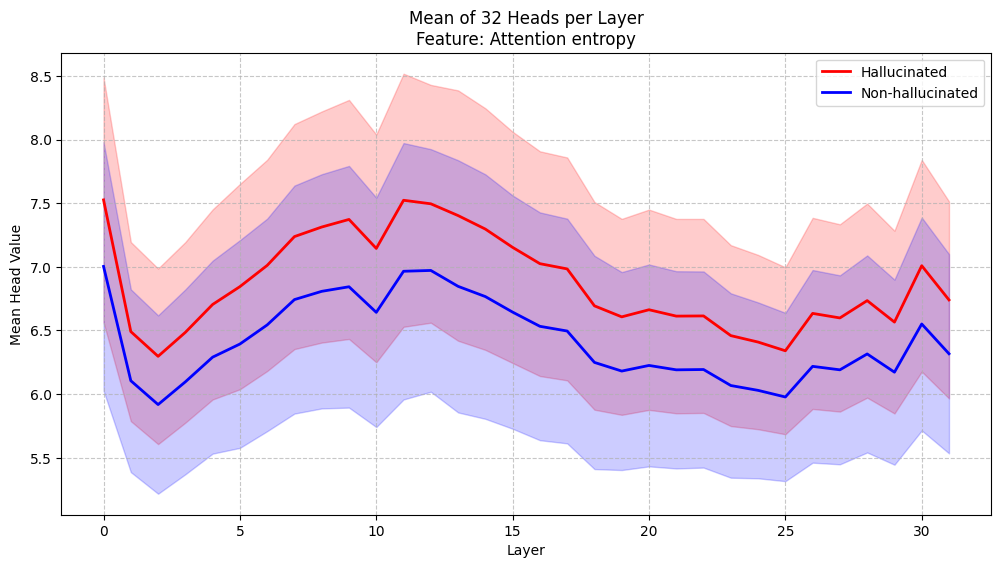}
    \caption{Attention entropy - LLaMA 3.1 8B Instruct}
    \label{fig:attention-entropy-LLaMA-single}
  \end{subfigure}
  \caption{Feature evolution across layers and comparison between hallucinated and non-
hallucinated response for LLaMA 3.1 8B Instruct.}
  \label{fig:LLaMA-pvals-single}
\end{figure}

We also analyzed how each feature evolves across layers over time by averaging attention heads, and compared these patterns for hallucinated responses. We compute the mean by taking the arithmetic average of a given feature across all attention heads in a given layer. As shown in Figure \ref{fig:LLaMA-pvals-single} for LLaMA 3.1 8B (and Appendix Figure \ref{fig:layer-wise} for Mistral 7B), both models exhibit higher average outgoingness scores and more diffuse attention distributions in hallucinated responses, suggesting impaired context sharing. In contrast, both Phi and Qwen show minimal or no variation in outgoingness across layers and no noticeable change in attention entropy (see Appendix, Figure \ref{fig:layer-wise}). Also, no clear pattern was observed that differentiates the earlier layers from the later ones in these behaviors.

\begin{figure}[h]
\includegraphics[width=0.48\textwidth]{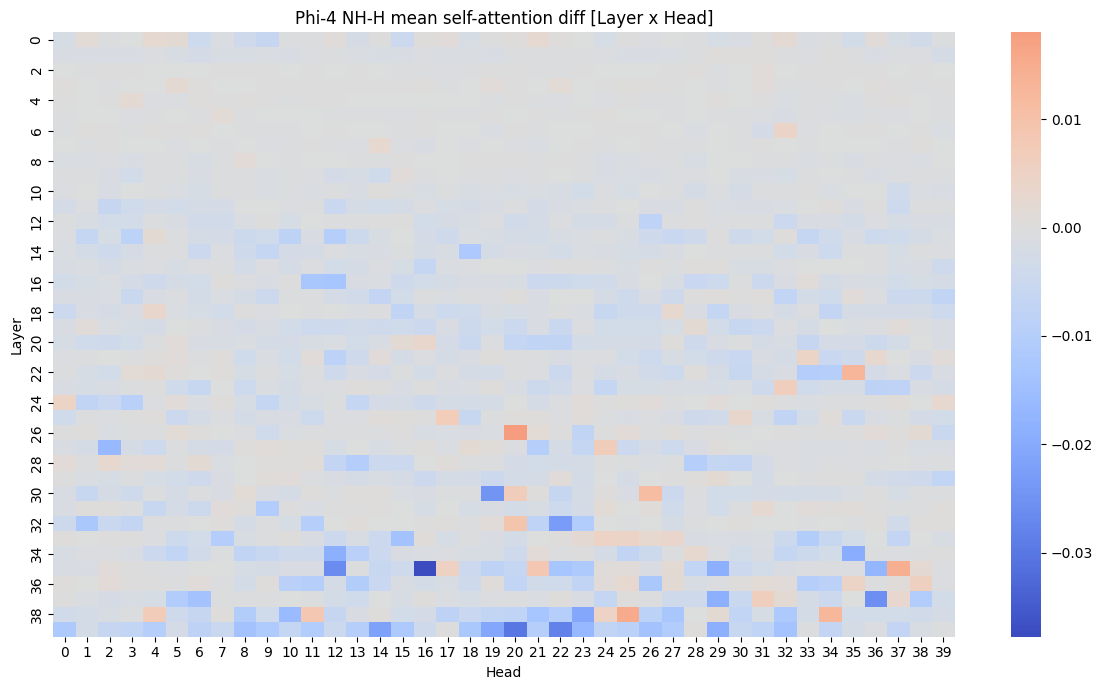}
\centering
\caption{Mean per-head self-attention differences between hallucinated and non-hallucinated responses for Phi-4.}
\label{fig: phi-self-attention}
\end{figure}

Excessive self-attention is frequently associated with a loss of contextual information from surrounding tokens, tracking closely with a reduction in the accuracy of transformer outputs \cite{arif2025paint}. Our experiments indicate that both Phi-4 and Qwen3-32B do not exhibit diffused attention, even when producing hallucinated content. To investigate this further, we analyzed the self-attention patterns across all layers and heads. We computed the average self-attention per head, then measured the difference in mean self-attention between hallucinated and non-hallucinated outputs. Figure \ref{fig: phi-self-attention} illustrates that for both models, in most attention heads, hallucinated responses correspond to higher self-attention—an effect that is especially evident in the later layers. Interestingly, this pattern is reversed and less statistically significant for the majority of heads in LLaMA and Mistral, where the hallucinated responses exhibit more diffuse attention (see Appendix, Figure \ref{fig:self-attention-differences}).

\subsection{Results}

We compare the results with Laplacian eigenvalues \cite{binkowski2025hallucination} as our main baseline due to its state-of-the-art performance, single-pass approach, and a simple yet effective and fast approach for identifying hallucination. 

We further compared our method with Eigen Score \cite{chen2024inside}, which detects hallucinations by generating multiple responses to a query and measuring the variance of the resulting information as a proxy for model uncertainty. In the original formulation, the user prompt is transformed into multiple template-based variants that preserve intent while inducing response diversity. However, this setup assumes access to prompt-engineering mechanisms that are not typically available in real-world use, where users submit a single prompt. Although prompt variants could be generated using an auxiliary LLM, doing so introduces additional computational cost and latency. To better reflect practical deployment constraints, we did not use template-based prompt reformulation. Instead, we generated multiple responses to the original prompt using a sampling temperature of 0.5, thereby inducing controlled stochasticity without modifying the input query.

We use a linear probe in three main feature configurations to distinguish hallucinated from non-hallucinated responses: (a) using the 10th, 50th, and 90th percentiles of outgoingness scores along with attention entropy; (b) using all deciles (10th through 90th) of outgoingness scores and attention entropy to assess the effect of including finer-grained percentiles; and (c) extending configuration (a) by adding the mean per-head self-attention scores to isolate the contribution of self-attention features to the probe’s performance. The best results were achieved using setting (c) (see Appendix, Table \ref{table:complete-results}).

The experiments indicate that the semi-local outgoingness scores and global attention entropy features yield substantial gains in area under the curve (AUC) compared to the Laplacian eigenvalues. When fine-grained outgoingness features were added, no significant gains were observed. When the average self-attention per-head values were added to the feature set, this increased AUC for all Phi-4 and Qwen3-23B models.

Qwen3-32B achieved the weakest performance among the evaluated models. A plausible explanation is its explicit multi-step reasoning mechanism. Since attention entropy and outgoingness are proxies for uncertainty and dispersion of information flow, they may be less sensitive to structured reasoning processes. As a result, these features have limited discriminative power across most attention heads.

\begin{figure}[h]
\includegraphics[width=0.5\textwidth]{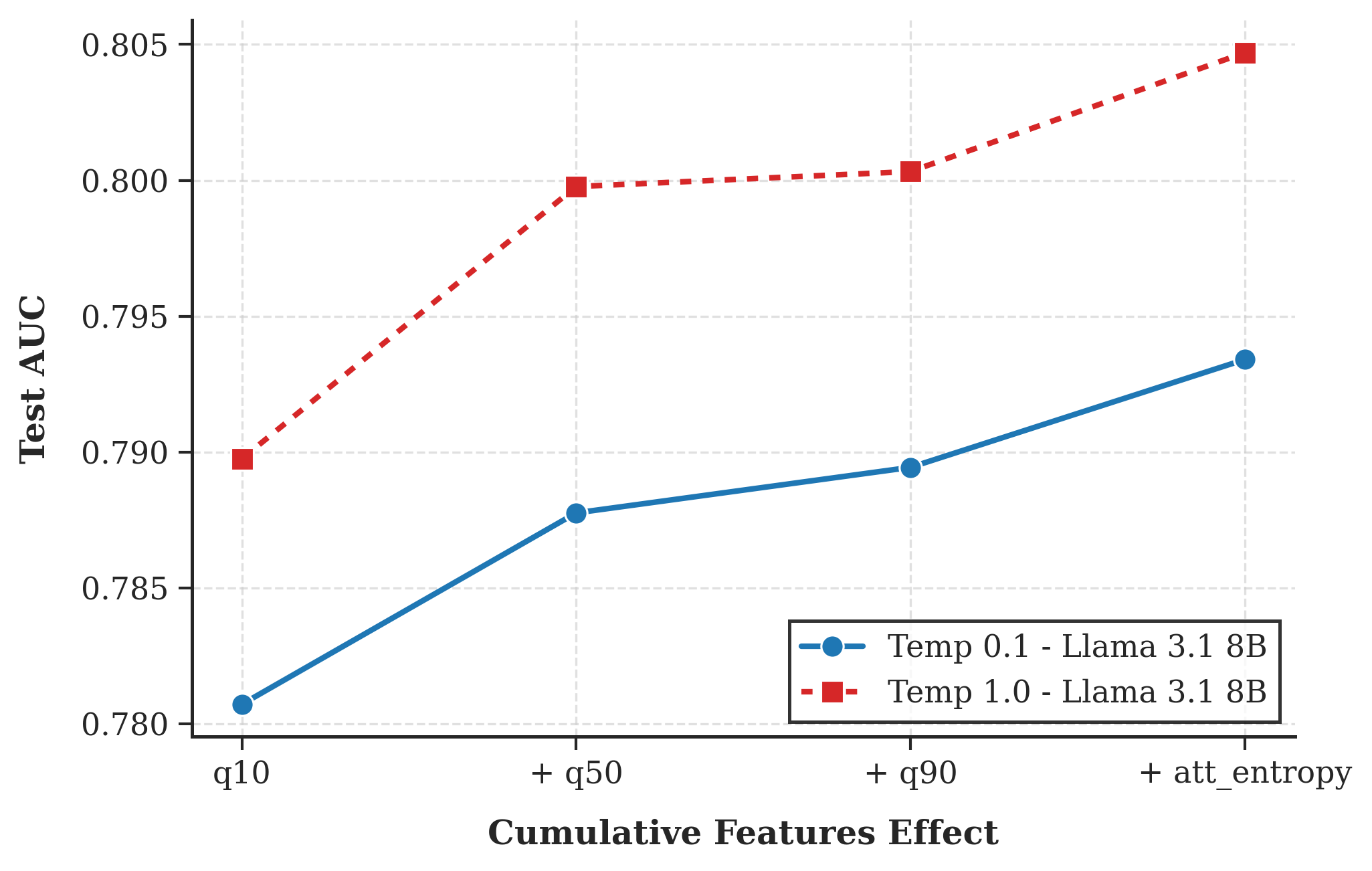}
\centering
\caption{Cumulative feature set effect for Llama 3.1 8B Instruct accross NQOPEN dataset.}
\label{fig: cumulative-evaluation}
\end{figure}

To assess the contribution of each feature type to hallucination detection, we evaluated the model's performance under a cumulative feature-addition scheme. As shown in Figure \ref{fig: cumulative-evaluation} and Table \ref{table:full-cumulative}, progressively adding feature groups generally led to consistent improvements in AUC. While incorporating the 90th percentile of outgoingness improved performance in most settings, its contribution was occasionally unstable, leading to marginal decreases in certain configurations. Overall, our method achieves an average improvement of approximately 3.4 AUC points over the selected baselines.

\begin{table*}[h]
\centering
\caption{AUC Comparison Between Laplacian eigenvalues, EigenScore, and our feature set classification results. Here we show the results for setting (a) which includes our base features. The best results however, were attained using setting (c).}
\begin{tabular}{|>{\centering\arraybackslash}m{1cm}|
                >{\centering\arraybackslash}m{2.2cm}|
                >{\centering\arraybackslash}m{4.5cm}|
                >{\centering\arraybackslash}m{2.0cm}|
                >{\centering\arraybackslash}m{2cm}|
                >{\centering\arraybackslash}m{2.2cm}|}
\hline
\multirow{2}{*}{\textbf{Temp}} & 
\multirow{2}{*}{\textbf{Benchmark}} & 
\multirow{2}{*}{\textbf{Model}} & 
\multicolumn{3}{c|}{\textbf{AUC}} \\
\cline{4-6}
& & & \textbf{LapEigvals} & \textbf{EigenScore} & \textbf{Ours - setting (a)} \\
\hline
0.1 & TruthfulQA & LLaMA 3.1 8B Instruct   & 80.52 & 83.79 & \textbf{85.34} \\
0.1 & TruthfulQA & Phi 4                  & 78.43 & 81.62 & \textbf{81.75} \\
0.1 & TruthfulQA & Mistral 7B Instruct v0.3& 78.80 & \textbf{81.69} & 80.00 \\
0.1 & TruthfulQA & Qwen3-32B & 77.36 & 72.17 & \textbf{80.17} \\
\hline
1.0 & TruthfulQA & LLaMA 3.1 8B Instruct   & 78.26 & 80.22 & \textbf{80.73} \\
1.0 & TruthfulQA & Phi 4                  & 74.46 & 79.82 & \textbf{84.00} \\
1.0 & TruthfulQA & Mistral 7B Instruct v0.3& 73.23 & 75.38 & \textbf{78.00} \\
1.0 & TruthfulQA & Qwen3-32B & \textbf{75.17} & 69.35 & 73.53 \\
\hline
0.1 & NQOPEN & LLaMA 3.1 8B Instruct & 75.67 & 71.49 & \textbf{79.34} \\
0.1 & NQOPEN & Phi 4 & 80.26 & 81.69 & \textbf{82.00} \\
0.1 & NQOPEN & Mistral 7B Instruct v0.3   & 77.10 & 79.45 & \textbf{79.75} \\
0.1 & NQOPEN & Qwen3-32B & 76.16 & 71.73 & \textbf{78.02} \\
\hline
1.0 & NQOPEN & LLaMA 3.1 8B Instruct  & 76.19 & 71.16 & \textbf{80.46} \\
1.0 & NQOPEN & Phi 4 & 83.61 & 84.11 & \textbf{84.49} \\
1.0 & NQOPEN & Mistral 7B Instruct v0.3   & 76.38 & \textbf{79.61} & 77.29 \\
1.0 & NQOPEN & Qwen3-32B & 78.49 & 73.63 & \textbf{81.73} \\
\hline
\end{tabular}
\label{table:partial-results}
\end{table*}

\begin{table*}[h]
\centering
\caption{Cumulative Feature Ablation (AUC \%) across TruthfulQA and NQOPEN datasets. Features are added sequentially.}
\begin{tabular}{|l|c|l|c|c|c|c|}
\hline
\textbf{Benchmark} & \textbf{Temp} & \textbf{Model} & \textbf{q10} & \textbf{+ q50} & \textbf{+ q90} & \textbf{+ att\_entropy} \\
\hline
\multirow{8}{*}{TruthfulQA} & \multirow{4}{*}{0.1} 
& LLaMA 3.1 8B Instruct      & 83.10 & 84.05 & 84.15 & 85.34 \\
& & Phi 4                      & 79.30 & 80.20 & 80.10 & 81.75 \\
& & Mistral 7B Instruct v0.3   & 77.50 & 78.20 & 78.35 & 80.00 \\
& & Qwen3-32B                   & 78.00 & 78.85 & 78.90 & 80.17 \\
\cline{2-7}
& \multirow{4}{*}{1.0} 
& LLaMA 3.1 8B Instruct      & 78.50 & 79.60 & 79.50 & 80.73 \\
& & Phi 4                      & 82.10 & 82.70 & 82.85 & 84.00 \\
& & Mistral 7B Instruct v0.3   & 76.10 & 76.85 & 76.70 & 78.00 \\
& & Qwen3-32B                  & 70.33 & 72.0 & 73.02 & 73.53 \\
\hline
\multirow{8}{*}{NQOPEN} & \multirow{4}{*}{0.1}
& LLaMA 3.1 8B Instruct      & 78.07 & 78.78 & 78.94 & 79.34 \\
& & Phi 4                      & 79.52 & 80.45 & 80.28 & 82.00 \\
& & Mistral 7B Instruct v0.3   & 77.28 & 77.50 & 78.17 & 79.75 \\
& & Qwen3-32B        & 74.18 & 74.98 & 77.35 & 78.02 \\
\cline{2-7}
& \multirow{4}{*}{1.0} 
& LLaMA 3.1 8B Instruct      & 78.98 & 79.98 & 80.03 & 80.46 \\
& & Phi 4                      & 82.88 & 82.99 & 83.02 & 84.49 \\
& & Mistral 7B Instruct v0.3   & 75.33 & 76.13 & 76.68 & 77.29 \\
& & Qwen3-32B        & 79.20 & 80.10 & 80.25 & 81.73 \\

\hline
\end{tabular}
\label{table:full-cumulative}
\end{table*}

\section{Related works}
Hallucination detection methods generally fall into two categories. The first one supposes that hallucination is the result of the model's sequence-level uncertainty or semantic inconsistency during token generation, and therefore its main focus is to identify hallucinated responses through measuring the amount of intravariability of multiple generated responses \cite{farquhar2024detecting}, \cite{chen2024inside}, \cite{ren2022out}. Such methods obtain multiple responses by applying top-p/top-k sampling strategy and then measure the semantic divergence among them. Although this approach has shown high detection rates, its performance depends on the number of generated samples and the variability of prompts with the same meaning. The more diversified the prompt and the higher the number of responses, the higher the detection accuracy \cite{chen2024inside}. This characteristic makes this approach costly, time-consuming, and therefore inadequate for real-time detection.

The second approach, which draws on more recent studies, has explored the use of specific attention-weight patterns to detect hallucinated responses in large language models (LLMs) \cite{binkowski2025hallucination} \cite{bazarova2025hallucination}. While it has demonstrated promising results, the studies that have followed the use of attention weights for hallucination detection often rely on simplified assumptions and lack a more in-depth analysis of the complex patterns co-occurring with hallucinations.

Bazarova et al. \cite{bazarova2025hallucination} posit that hallucinated responses introduce a unique and measurable topological dissimilarity in the attention graphs connecting the response to the prompt. However, this framework may oversimplify the nature of attentional failures. It does not fully account for scenarios such as information bottlenecks, strong attentional links between response tokens and irrelevant prompt tokens, or the diverse patterns that may emerge across different domains and LLM architectures.

Similarly, Binkowski et al. \cite{binkowski2025hallucination} hypothesize that hallucinations are a direct consequence of information bottlenecks and propose using the eigenvalues of the Laplacian matrix of attention graphs to detect them. While innovative, Laplacian eigenvalues reflect a range of graph characteristics beyond bottlenecks, including graph bipartiteness and the number of spanning trees, which may or may not be related to hallucination. Despite their hypothesis, the authors did not conduct empirical experiments to show how Laplacian eigenvalues vary depending on whether a response is hallucinated. Furthermore, this method provides a global perspective on graph properties, potentially obscuring localized attentional features that could be more directly associated with hallucinated responses. Moreover, the authors show that the higher the number of eigenvalues, the better the distinguishability of their feature set and report that their best results were achieved when 100 eigenvalues were extracted from each attention head. However, this puts a hard constraint on the minimum response length, making the method less effective for short responses.

In the current study, we carried out a series of analyses to examine the relationship between attention graph bottlenecks and hallucinations, mapping out how these patterns manifest across different LLM architectures. Afterward, we propose a set of features that enable high-performance hallucination detection, and explain why and how hallucination occurs across different LLMs.

\section{Limitations}

\paragraph{Theoretical and Operational Confounders.} 
Our topological framework cannot distinctly separate pathological context-sharing impairments from either absolute parametric knowledge gaps or functional attention sinks (\eg, initial tokens or punctuation). Because absolute data omissions and architectural attention anchors both induce localized distortions in attention distributions, the observed signatures risk confounding true semantic bottlenecks with benign structural properties or generic model uncertainty.

\paragraph{Architectural and Evaluation Scope.} 
The proposed method exhibits explicit architectural dependencies, resulting in significant performance drops on reasoning-dense models like Qwen3-32B. A recent study \cite{korbak2025chain} has suggested that reasoning traces can contain signals associated with incorrect, deceptive, or fabricated information in the final response; however, our analysis does not attempt to establish a causal relationship between reasoning mechanisms and the proposed topological features. Rather, our objective was to examine whether the identified information-flow patterns remain observable in a reasoning-oriented model or whether they differ under such generation processes. Furthermore, our empirical validation is restricted to single-turn, English short-form question-answering which is the predominant setting adopted by existing hallucination detection benchmarks. The generalizability of these topological indices to non-English syntax, long-form multi-turn generation, or complex agentic reasoning pathways remains unverified.


\section{Conclusion}
In this study, we conducted a comprehensive set of analyses and provided evidence that hallucinations manifest differently across various large language models (LLMs). We demonstrate that simplified attention-based properties alone are insufficient to characterize these phenomena. To address this shortcoming, we introduced a set of response-length-independent features to capture a broader range of hallucination patterns. Through a detailed examination and extensive experimentation, we observed that while attention behaviors during hallucination vary across different LLMs, impaired context sharing is a consistent topological signature correlated with hallucinations across all models examined. 

Subsequently, using the proposed features, a simple linear probe achieved competitive hallucination-detection performance and consistently outperformed the selected baselines across the evaluated models, benchmarks, and generation settings. Our selection of diverse LLMs allowed us to analyze both shared and distinct behavioral patterns. However, the impact of model size, specifically the number of parameters, was not investigated in this work. For instance, although the Phi-4 model, which has approximately twice as many parameters as Llama 3.1 8B Instruct and Mistral 7B Instruct, exhibits notably different behavior, it remains unclear whether these differences are attributable to model size or other architectural factors. Therefore, the interplay between model size and attention behavior during hallucination can be investigated further in future work. Also, our results indicate that producing correct responses requires tokens to balance reliance on their own semantic representations with the contextual information from other tokens. Hallucinations can therefore be interpreted as a failure to correctly integrate contextual information from the prompt. Still, it remains an open question how internal model dynamics differ when hallucination is caused by genuine knowledge gaps rather than misaligned context sharing.

\bibliographystyle{IEEEtran}
\bibliography{custom}
\appendix

\section{Additional Experiments}

\subsection{Complementary results}

\begin{figure*}[h]
  \centering
  
  \begin{subfigure}[b]{0.6\textwidth}
    \includegraphics[width=\linewidth]{img/ricci-diff-LLaMA.png}
    \caption{LLaMA 3.1 8B Instruct}
    \label{fig:ricci-LLaMA}
  \end{subfigure}
  \hfill
  \begin{subfigure}[b]{0.6\textwidth}
    \includegraphics[width=\linewidth]{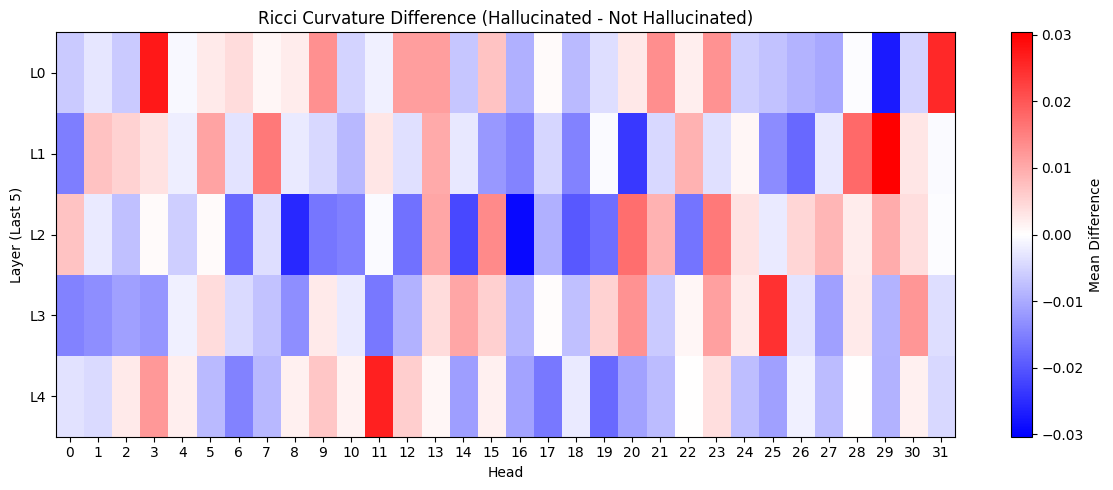}
    \caption{Mistral 7B Instruct}
    \label{fig:ricci-misteral}
  \end{subfigure}
  \begin{subfigure}[b]{0.6\textwidth}
    \includegraphics[width=\linewidth]{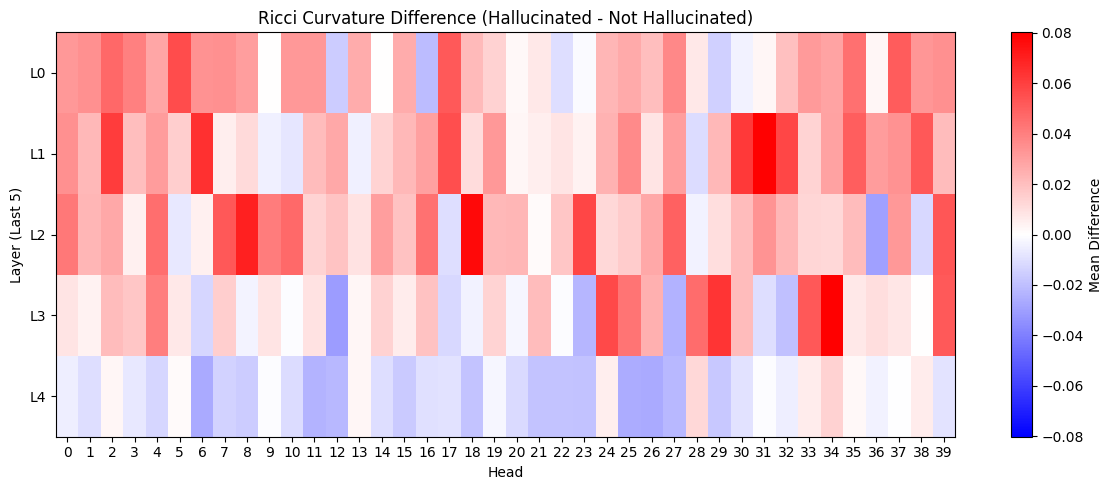}
    \caption{Phi-4}
    \label{fig:ricci-phi}
  \end{subfigure}
    \begin{subfigure}[b]{0.8\textwidth}
    \includegraphics[width=\linewidth]{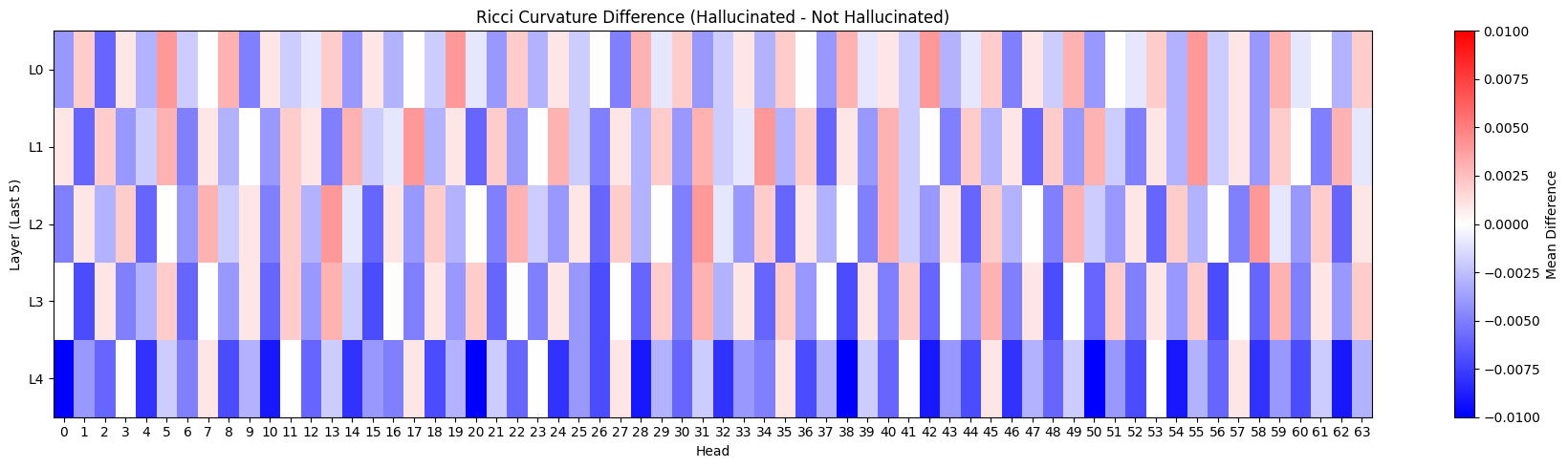}
    \caption{Qwen3-32B}
    \label{fig:ricci-qwen}
  \end{subfigure}

  \caption{The difference between the mean values of Forman-Ricci curvatures per head for hallucinated and not hallucinated responses. For the majority of the attention heads, hallucinated responses have a lower curvature values which is an indication of a less fluid information flow. This pattern is more evident accross all layers in LLaMA and Mistral. For Phi and Qwen, the last layer presents less fluid context sharing for hallucinated responses.}
  \label{fig:ricci-differences}
\end{figure*}

\begin{figure*}[t]
  \centering
  
  \begin{subfigure}[b]{0.7\textwidth}
    \includegraphics[width=\linewidth]{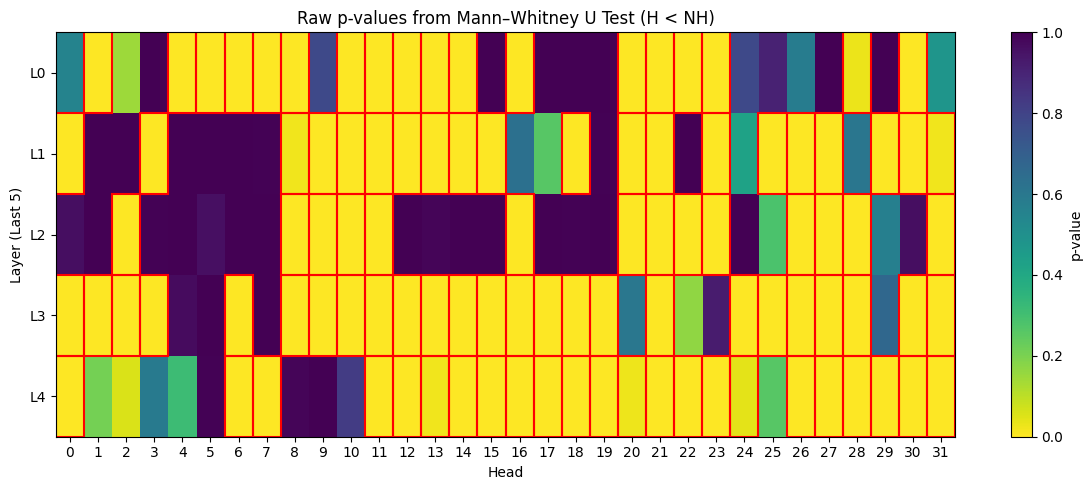}   
    \caption{LLaMA 3.1 8B Instruct}
    \label{fig:ricci-pval-LLaMA}
  \end{subfigure}
  \hfill
  \begin{subfigure}[b]{0.7\textwidth}
    
    \includegraphics[width=\linewidth]{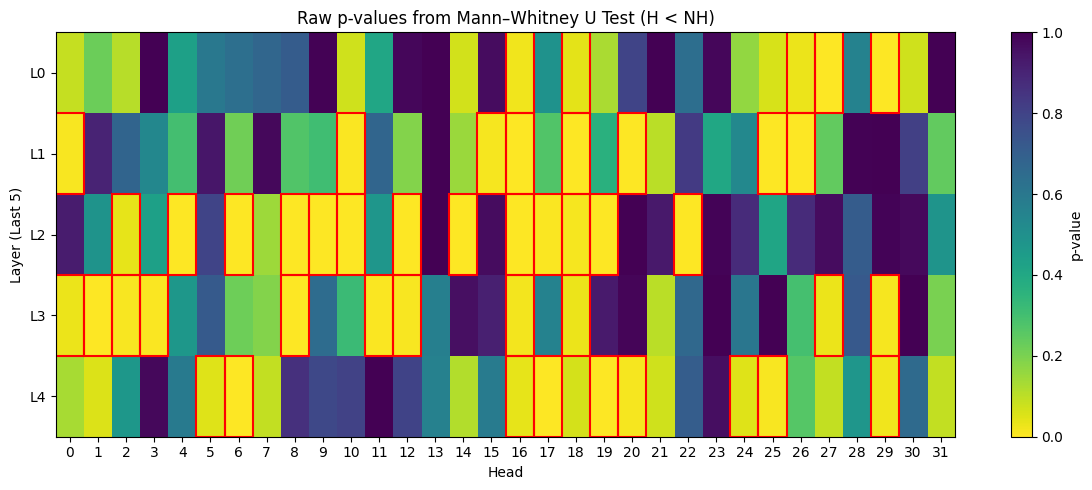}
    \caption{Mistral 7B Instruct}
    \label{fig:ricci-misteral}
  \end{subfigure}
  \begin{subfigure}[b]{0.7\textwidth}
    \includegraphics[width=\linewidth]{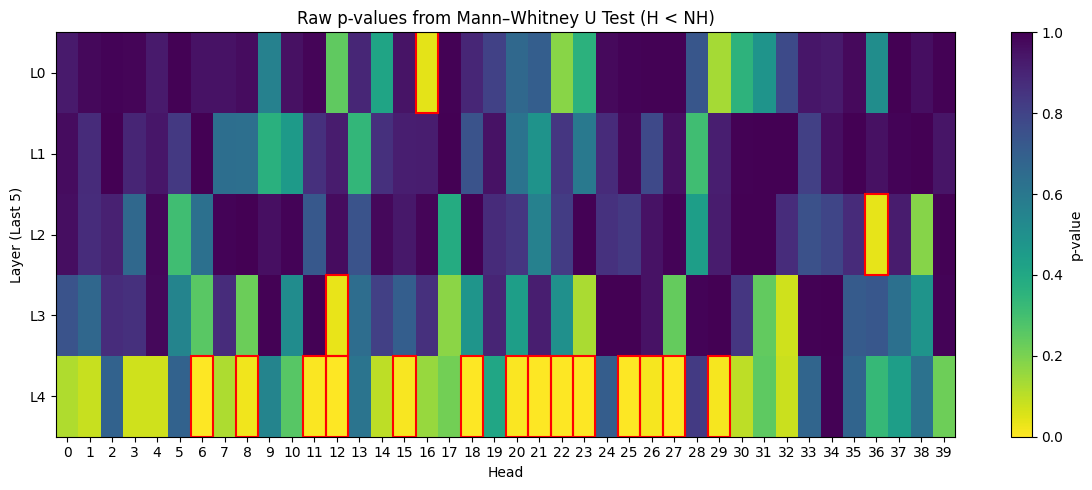}
    \caption{Phi-4}
    \label{fig:ricci-phi}
  \end{subfigure}
    \begin{subfigure}[b]{0.9\textwidth}
    \includegraphics[width=\linewidth]{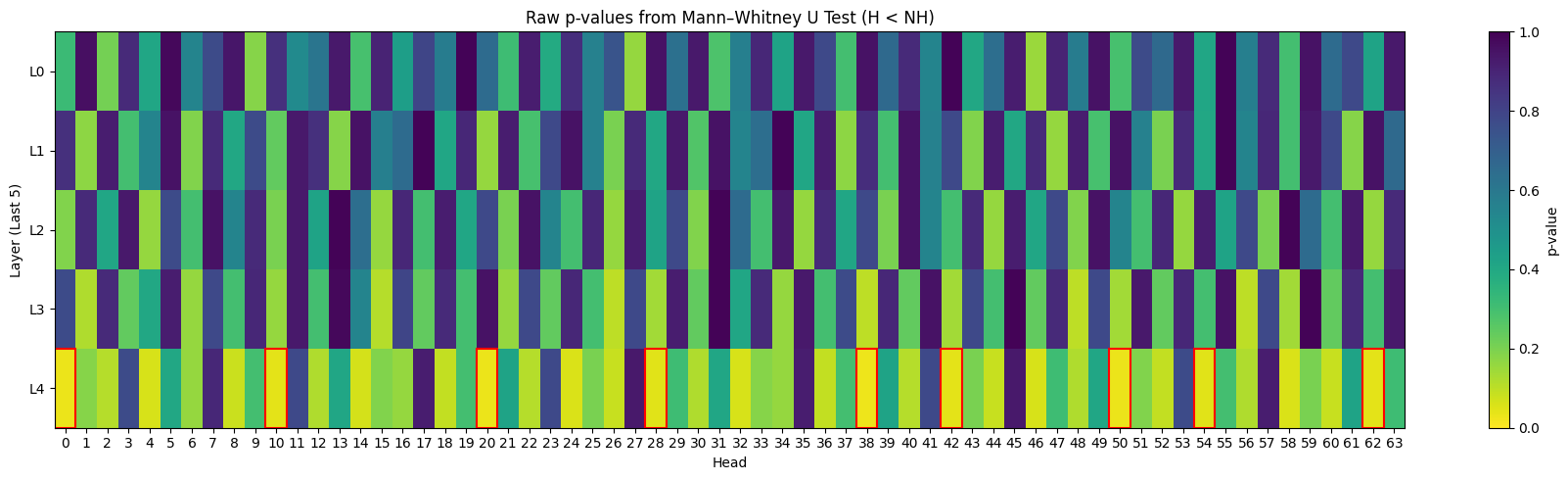}
    \caption{Qwen3-32B}
    \label{fig:ricci-qwen}
  \end{subfigure}

  \caption{The p-values of the differences between the mean values of Forman-Ricci curvatures per head for hallucinated and not hallucinated responses. The heads with statistical signiuficance are highlited with red lines.}
  \label{fig:p-values-ricci-differences}
\end{figure*}

\begin{figure*}[t]
  \centering

  \begin{subfigure}[b]{0.30\textwidth}
    \centering
    \includegraphics[width=\linewidth]{img/10th-percentile-LLaMA.png}
    \caption{Outgoingness -- LLaMA}
    \label{fig:app-outgoingness-llama}
  \end{subfigure}
  \hfill
  \begin{subfigure}[b]{0.30\textwidth}
    \centering
    \includegraphics[width=\linewidth]{img/attention-entropy-LLaMA.png}
    \caption{Attention entropy -- LLaMA}
    \label{fig:app-entropy-llama}
  \end{subfigure}
  \hfill
  \begin{subfigure}[b]{0.30\textwidth}
    \centering
    \includegraphics[width=\linewidth]{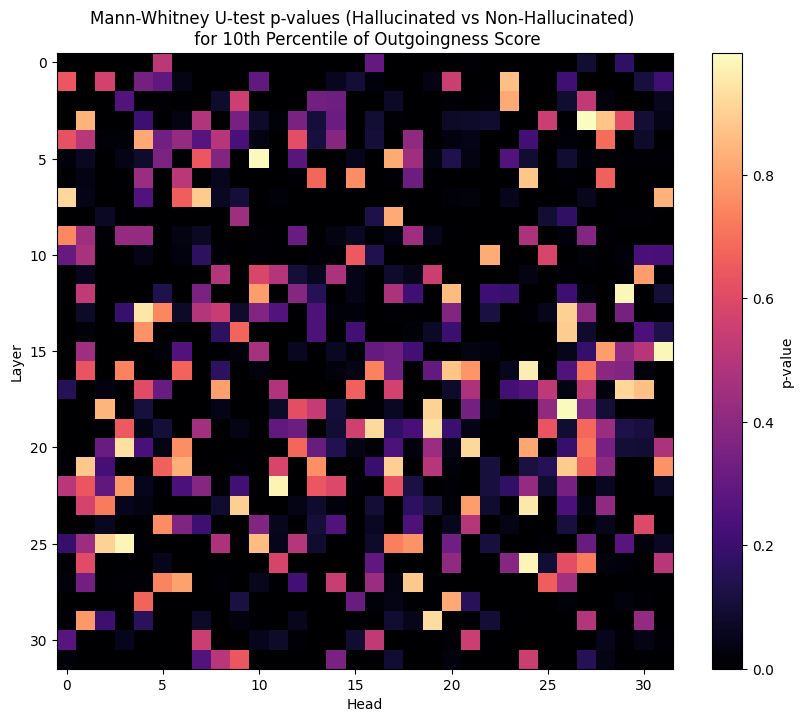}
    \caption{Outgoingness -- Mistral}
    \label{fig:app-outgoingness-mistral}
  \end{subfigure}

  \vspace{2mm}

  \begin{subfigure}[b]{0.30\textwidth}
    \centering
    \includegraphics[width=\linewidth]{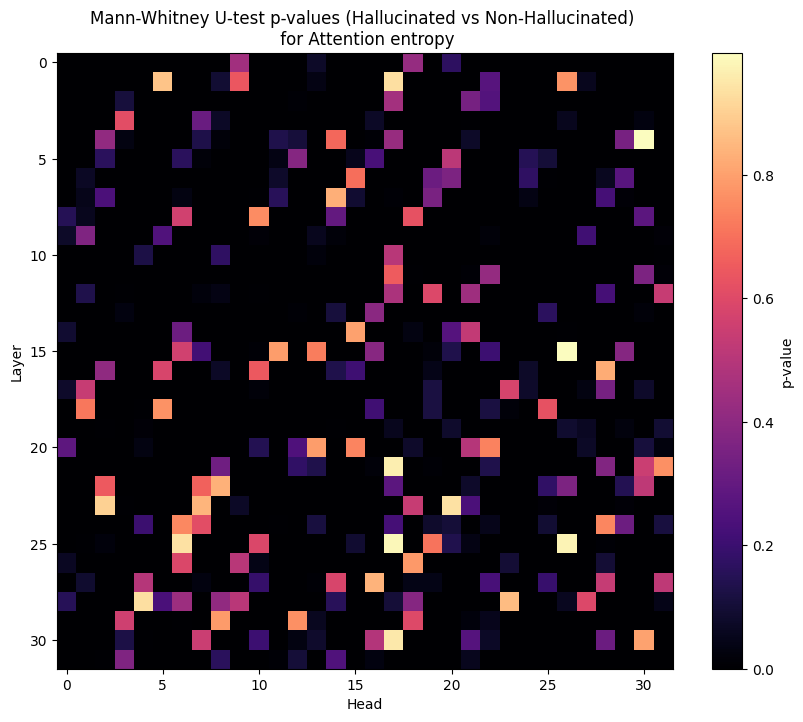}
    \caption{Attention entropy -- Mistral}
    \label{fig:app-entropy-mistral}
  \end{subfigure}
  \hfill
  \begin{subfigure}[b]{0.30\textwidth}
    \centering
    \includegraphics[width=\linewidth]{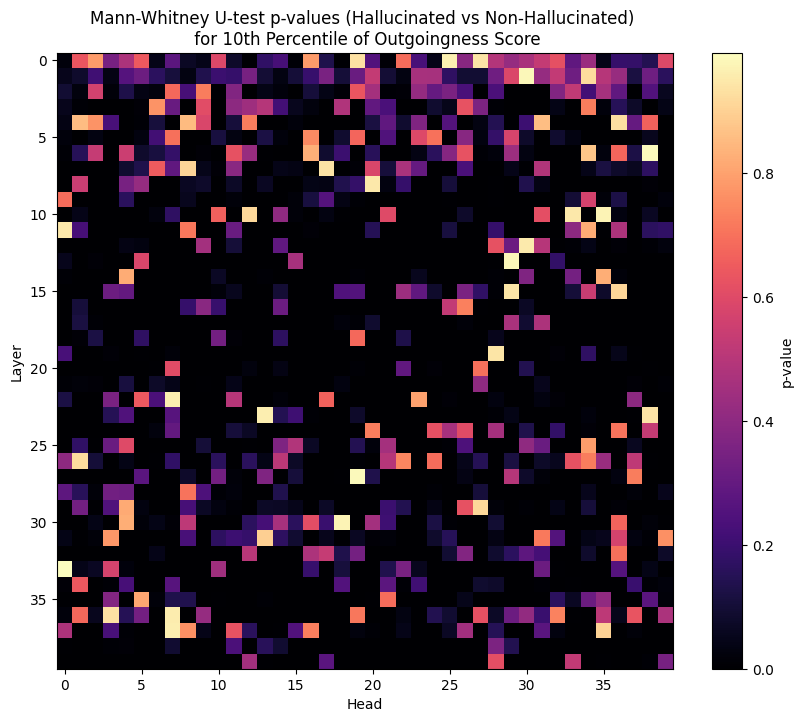}
    \caption{Outgoingness -- Phi-4}
    \label{fig:app-outgoingness-phi}
  \end{subfigure}
  \hfill
  \begin{subfigure}[b]{0.30\textwidth}
    \centering
    \includegraphics[width=\linewidth]{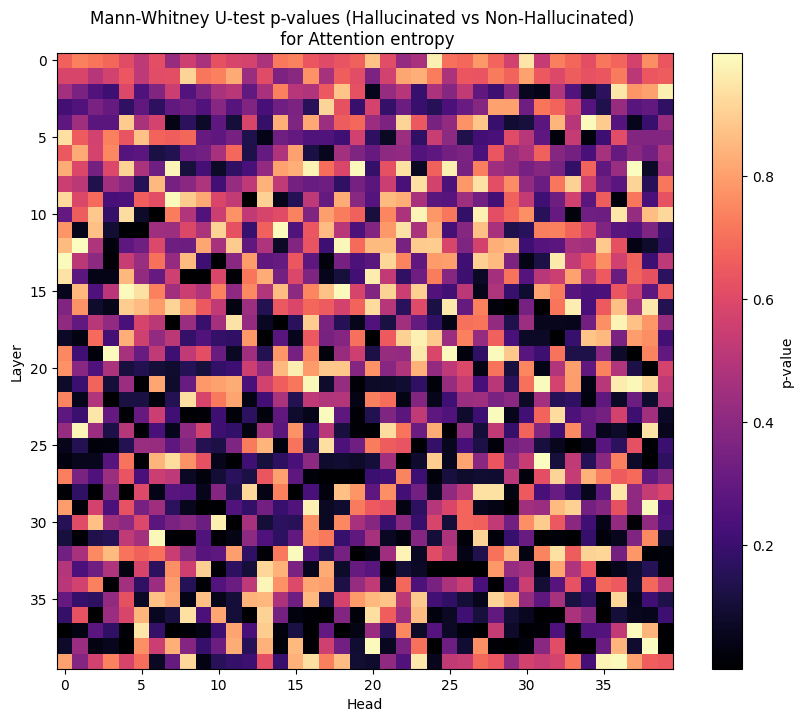}
    \caption{Attention entropy -- Phi-4}
    \label{fig:app-entropy-phi}
  \end{subfigure}

  \vspace{2mm}

  \begin{subfigure}[b]{0.30\textwidth}
    \centering
    \includegraphics[width=\linewidth]{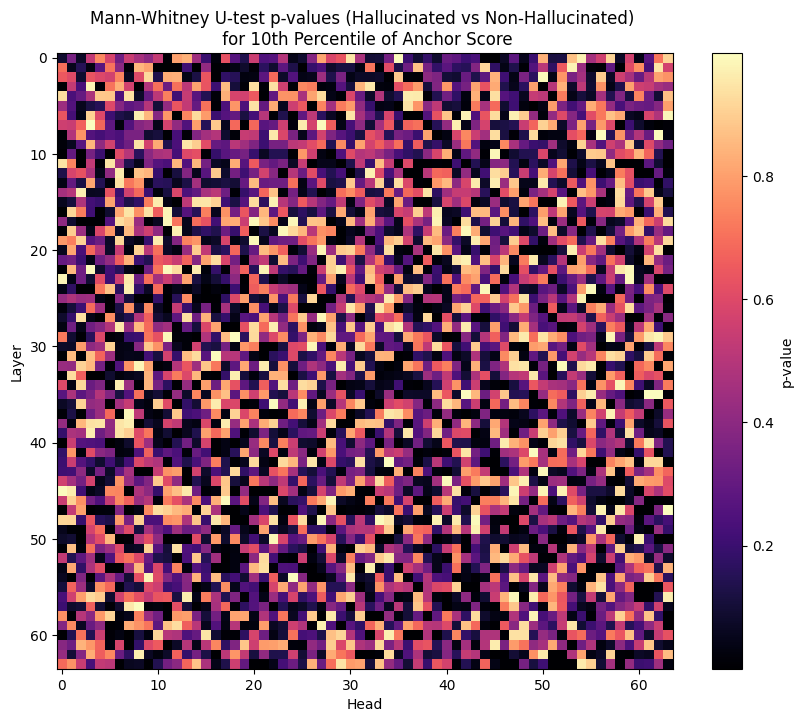}
    \caption{Outgoingness -- Qwen3-32B}
    \label{fig:app-outgoingness-qwen}
  \end{subfigure}
  \hspace{0.04\textwidth}
  \begin{subfigure}[b]{0.30\textwidth}
    \centering
    \includegraphics[width=\linewidth]{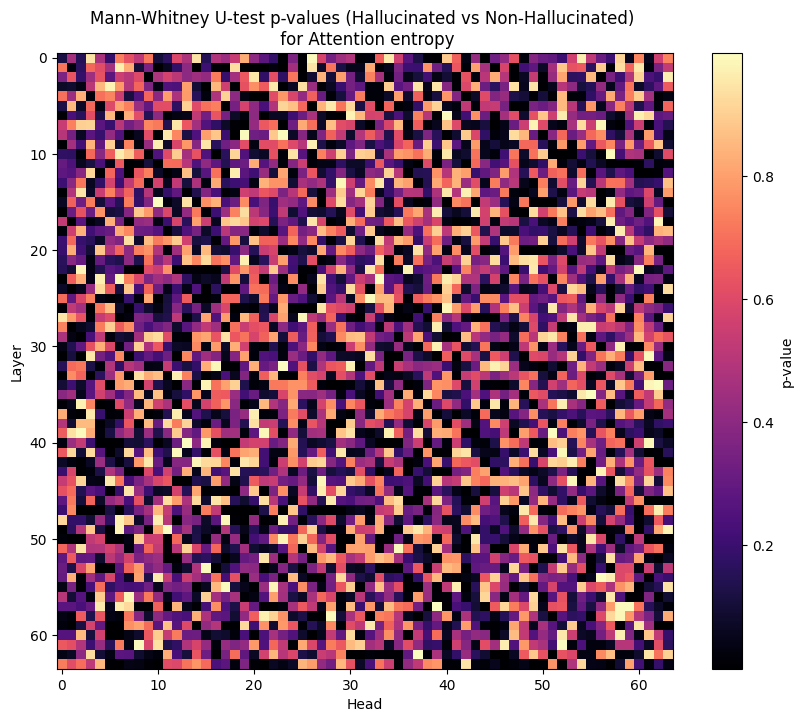}
    \caption{Attention entropy -- Qwen3-32B}
    \label{fig:app-entropy-qwen}
  \end{subfigure}

  \caption{P-values for the per-head comparisons of the 10th-percentile outgoingness scores and attention entropy across all layers. LLaMA 3.1 8B Instruct exhibits the strongest separation across the majority of attention heads.}
  \label{fig:features-pvalues-part1}
\end{figure*}

\begin{figure*}[h]
  \centering
  \begin{subfigure}[b]{0.48\textwidth}
    \includegraphics[width=\linewidth]{img/LLaMA-10th-percentile-layer-wise.png}
    \caption{Outgoingness - LLaMA 3.1 8B Instruct}
    \label{fig:10th-percentile-LLaMA}
  \end{subfigure}
  \begin{subfigure}[b]{0.48\textwidth}
    \includegraphics[width=\linewidth]{img/LLaMA-attention-entropy-layer-wise.png}
    \caption{Attention entropy - LLaMA 3.1 8B Instruct}
    \label{fig:attention-entropy-LLaMA}
  \end{subfigure}
  
  \begin{subfigure}[b]{0.48\textwidth}
    \includegraphics[width=\linewidth]{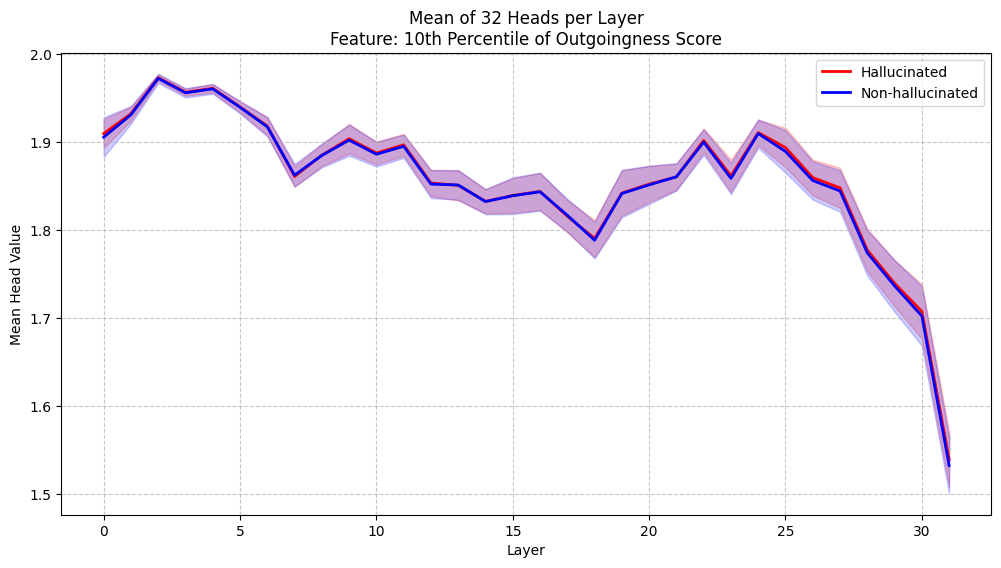}
    \caption{Outgoingness - Mistral 7B Instruct}
    \label{fig:10th-percentile-misteral}
  \end{subfigure}
  \begin{subfigure}[b]{0.48\textwidth}
    \includegraphics[width=\linewidth]{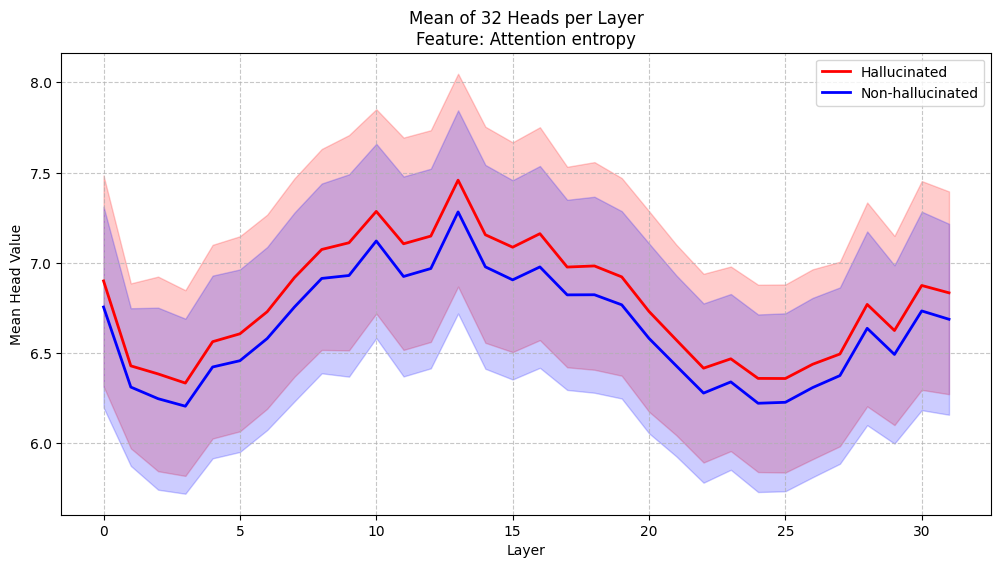}
    \caption{Attention entropy - Mistral 7B Instruct}
    \label{fig:attention-entropy-misteral}
  \end{subfigure}
  
  \begin{subfigure}[b]{0.48\textwidth}
    \includegraphics[width=\linewidth]{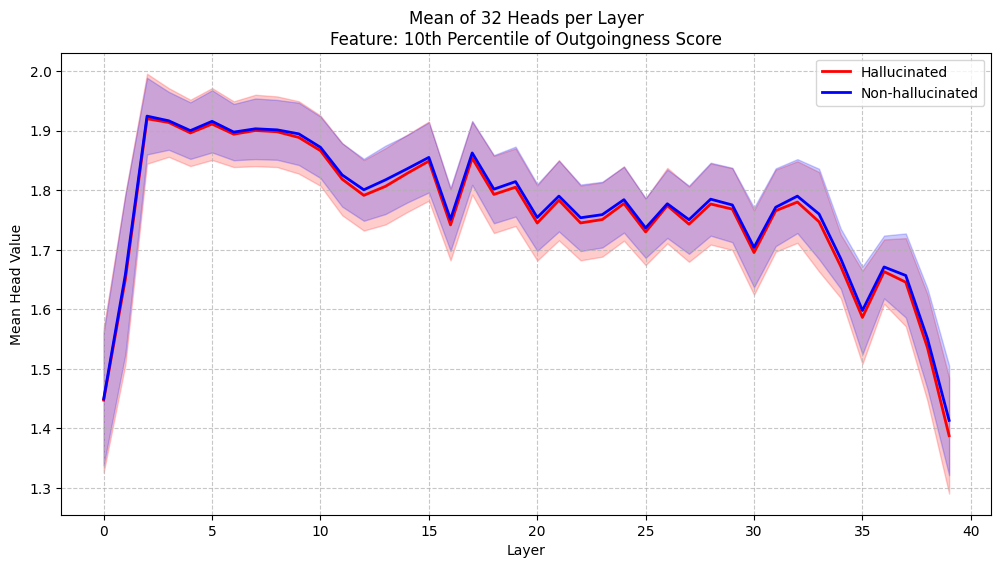}
    \caption{Outgoingness - Phi-4}
    \label{fig:10th-percentile-phi}
  \end{subfigure}
  \begin{subfigure}[b]{0.48\textwidth}
    \includegraphics[width=\linewidth]{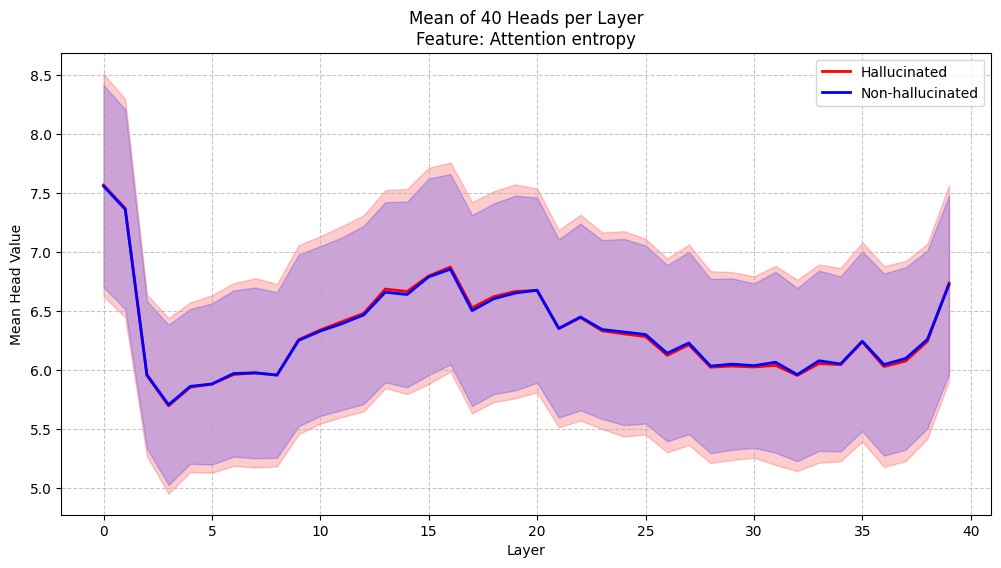}
    \caption{Attention entropy - Phi-4}
    \label{fig:attention-entropy-phi}
  \end{subfigure}
    \begin{subfigure}[b]{0.48\textwidth}
    \includegraphics[width=\linewidth]{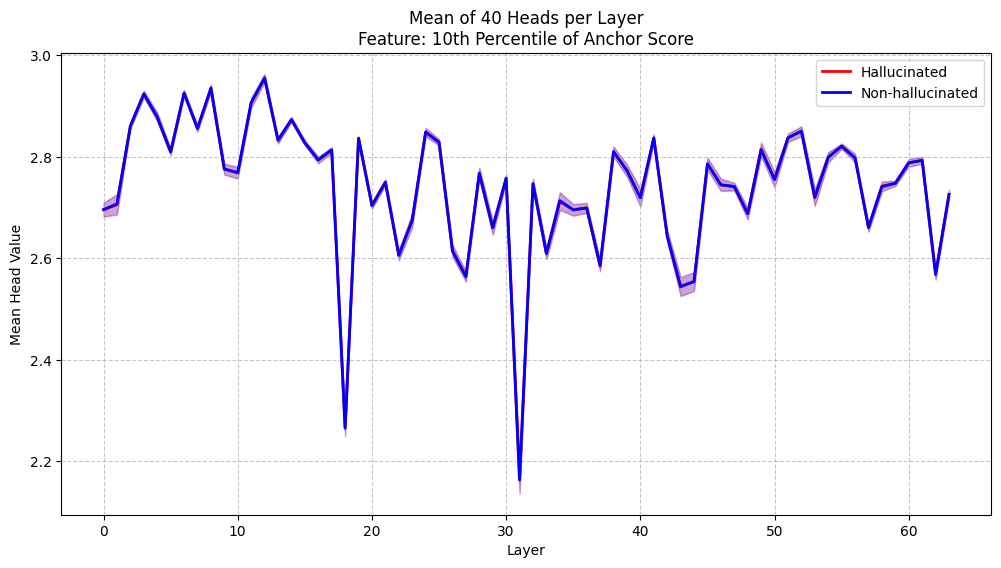}
    \caption{Outgoingness - Qwen3-32B}
    \label{fig:10th-percentile-phi}
  \end{subfigure}
  \begin{subfigure}[b]{0.48\textwidth}
    \includegraphics[width=\linewidth]{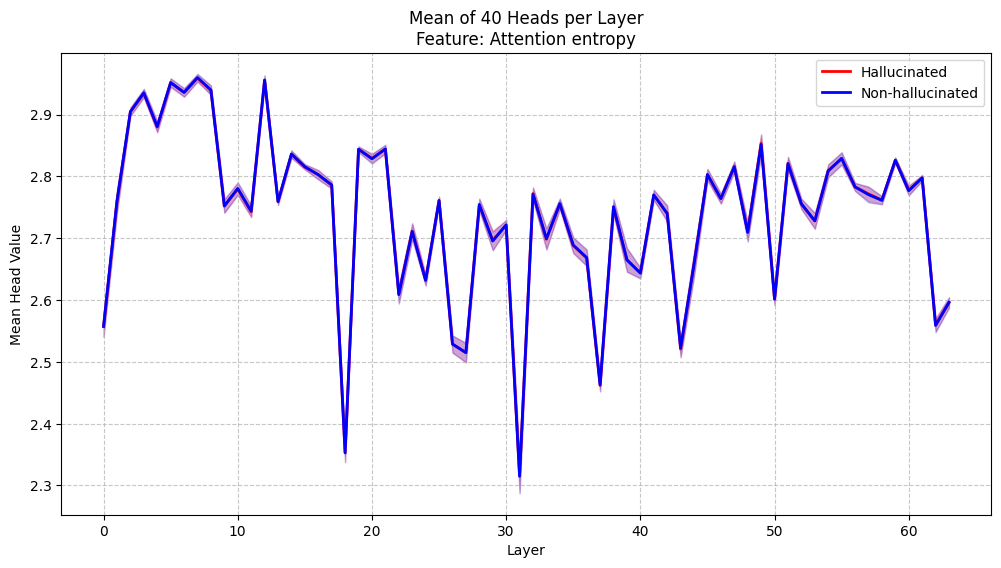}
    \caption{Attention entropy - Qwen3-32B}
    \label{fig:attention-entropy-phi}
  \end{subfigure}
\caption{Feature average evolution across layers and comparison between hallucinated and non-hallucinated responses. Both LLaMA 3.1 Instruct and Mistral 7B Instruct shows diffused attention whereas for Phi-4 shows little to no noticeable difference in outgoingness or attention distribution between the two response types. Also, for LLaMA 3.1 Instruct and Mistral 7B Instruct, the average of 10th percentile of the outgoingness score over attention heads is higher for hallucinated response in the terminal layers. The same pattern was observed for 50th and 90th percentiles.}
\label{fig:layer-wise}
\end{figure*}

\begin{figure*}[t]
  \centering

  \begin{subfigure}[b]{0.48\textwidth}
    \centering
    \includegraphics[width=\linewidth]{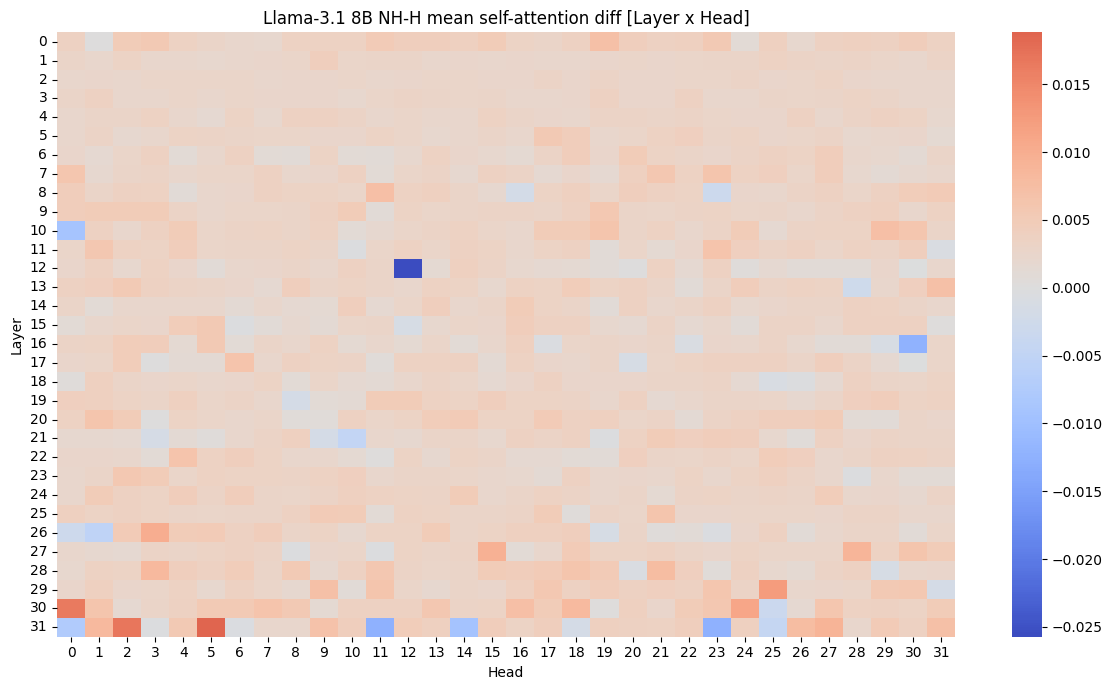}
    \caption{LLaMA 3.1 8B Instruct}
    \label{fig:app-self-attention-llama}
  \end{subfigure}
  \hfill
  \begin{subfigure}[b]{0.48\textwidth}
    \centering
    \includegraphics[width=\linewidth]{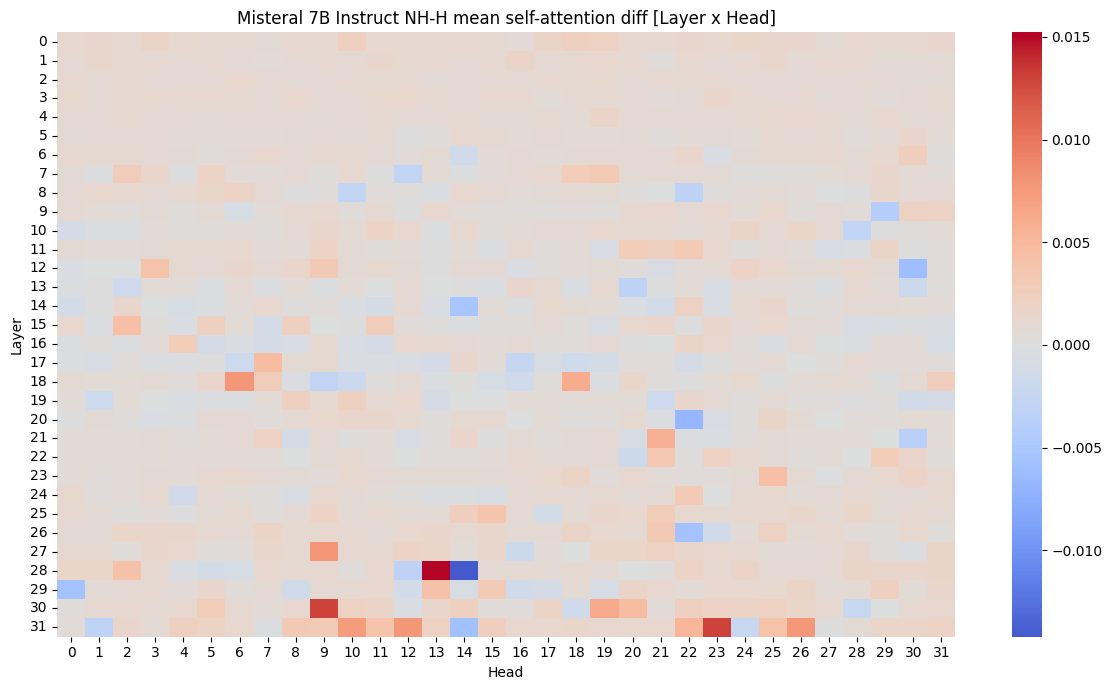}
    \caption{Mistral 7B Instruct}
    \label{fig:app-self-attention-mistral}
  \end{subfigure}

  \vspace{2mm}

  \begin{subfigure}[b]{0.48\textwidth}
    \centering
    \includegraphics[width=\linewidth]{img/self-attention-phi4.png}
    \caption{Phi-4}
    \label{fig:app-self-attention-phi}
  \end{subfigure}
  \hfill
  \begin{subfigure}[b]{0.48\textwidth}
    \centering
    \includegraphics[width=\linewidth]{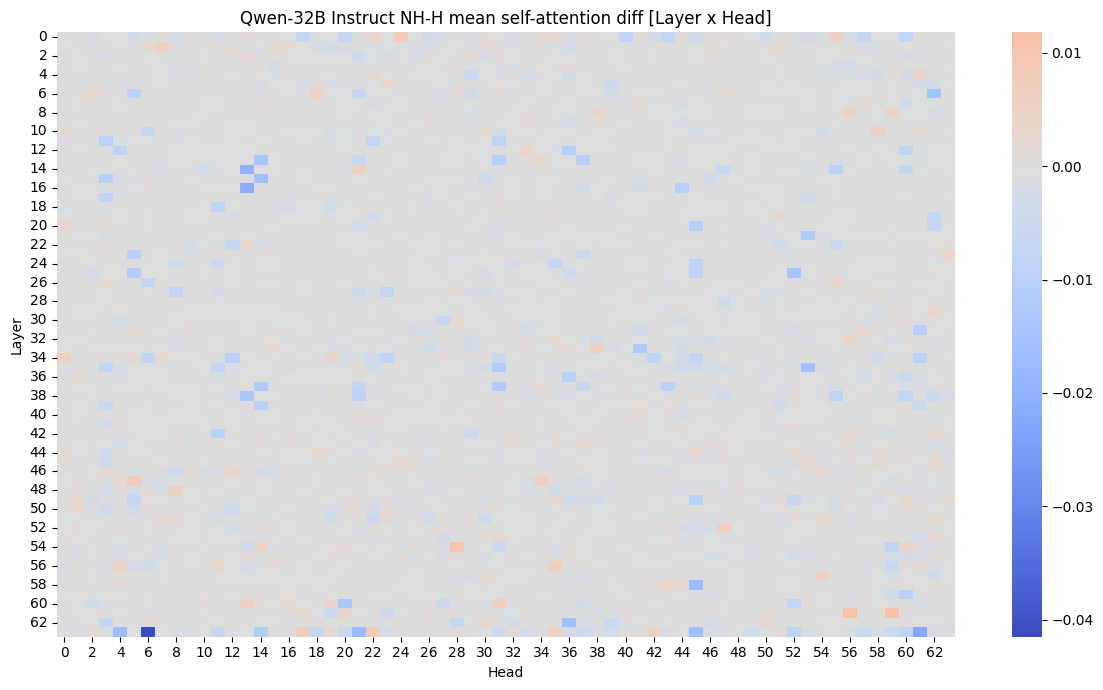}
    \caption{Qwen3-32B}
    \label{fig:app-self-attention-qwen}
  \end{subfigure}

  \caption{
  Mean per-head self-attention differences between hallucinated and
  non-hallucinated responses across the evaluated models.
  Unlike LLaMA 3.1 8B Instruct and Mistral 7B Instruct, Phi-4 and
  Qwen3-32B exhibit higher self-attention for hallucinated responses
  across a large proportion of layers and attention heads.
  }
  \label{fig:self-attention-differences}
\end{figure*}

\clearpage
\begin{sidewaystable}[p]
\centering
\begin{tabular}{|>{\centering\arraybackslash}m{1cm}|
                >{\centering\arraybackslash}m{2.5cm}|
                >{\centering\arraybackslash}m{4.5cm}|
                >{\centering\arraybackslash}m{2.2cm}|
                >{\centering\arraybackslash}m{2.2cm}|
                >{\centering\arraybackslash}m{2.2cm}|
                >{\centering\arraybackslash}m{2.2cm}|
                >{\centering\arraybackslash}m{2.2cm}|}
\hline
\multirow{2}{*}{\textbf{Temp}} & 
\multirow{2}{*}{\textbf{Benchmark}} &
\multirow{2}{*}{\textbf{Model}} & 
\multicolumn{5}{c|}{\textbf{AUC}} \\
\cline{4-8}
& & & \textbf{LapEigvals} & \textbf{EigenScore} & 
\makecell{\textbf{Ours - } \\ \textbf{setting (a)}} & 
\makecell{\textbf{Ours - } \\ \textbf{setting (b)}} & 
\makecell{\textbf{Ours - } \\ \textbf{setting (c)}} \\
\hline

0.1 & TruthfulQA & LLaMA 3.1 8B Instruct & 80.52 & 83.79 & 85.34 & \textbf{85.56} & 85.47\\
0.1 & TruthfulQA & Phi 4                  & 78.43 & 81.62 & 81.75 & 81.80 & \textbf{82.41}\\
0.1 & TruthfulQA & Mistral 7B Instruct v03 & 78.80 & \textbf{81.69} & 80.00 & 80.17 & 80.11\\
0.1 & TruthfulQA & Qwen3-32B & 77.36 & 72.17 & 80.17 & 80.33 & \textbf{80.47} \\
\hline
1.0 & TruthfulQA & LLaMA 3.1 8B Instruct & 78.26 & 80.22 & 80.73 & 80.25 & \textbf{81.00}\\
1.0 & TruthfulQA & Phi 4                  & 74.46 & 79.82 & 84.00 & 83.98 & \textbf{85.31} \\
1.0 & TruthfulQA & Mistral 7B Instruct v03 & 73.23 & 75.38 & 78.00 & \textbf{78.68} & 78.19\\
1.0 & TruthfulQA & Qwen3-32B & \textbf{75.17} & 69.35 & 73.53 & 74.11 & 74.65 \\
\hline
0.1 & NQOPEN & LLaMA 3.1 8B Instruct  & 75.67 & 71.49 & 79.34 & \textbf{79.38} & 79.33 \\
0.1 & NQOPEN & Phi 4                      & 80.26 & 81.69 & 82.00 & 82.36 & \textbf{83.34}\\
0.1 & NQOPEN & Mistral 7B Instruct v03 & 77.10 & 79.45 & 79.75 & 79.34 & \textbf{79.77}\\
0.1 & NQOPEN & Qwen3-32B & 76.16 & 71.73 & 78.02 & 78.16 & \textbf{78.37}  \\
\hline
1.0 & NQOPEN & LLaMA 3.1 8B Instruct  & 76.19 & 71.16 & \textbf{80.46} & 80.15 & 80.21\\
1.0 & NQOPEN & Phi 4                      & 83.61 & 84.11 & 84.49 & 84.77 & \textbf{85.12} \\
1.0 & NQOPEN & Mistral 7B Instruct v03 & 76.38 & \textbf{79.61} & 77.29 & 77.92 & 77.13\\
1.0 & NQOPEN & Qwen3-32B & 78.49 & 73.63 & 81.73  & 81.43  & \textbf{81.79} \\
\hline
\end{tabular}
\caption{AUC comparison between LapEigvals, EigenScore, and our method across three feature settings. Setting (a) includes the 10th, 50th, and 90th percentile outgoingness per-head features, along with attention entropy. Setting (b) extends (a) by incorporating all deciles (10th–90th) to evaluate the impact of finer-grained local information. Setting (c) augments (a) with average per-head self-attention values as additional features.}
\label{table:complete-results}
\end{sidewaystable}

\clearpage
\subsection{LLM-as-judge}
To assess the reliability of the LLM-as-judge evaluations, we randomly sampled 100 prompts, established the correct answer for each prompt, and manually evaluated the judgments produced by GPT-4. Although none of the evaluated model responses exhibited severe label imbalance (moderate average ration of 15:85 for incorrect and correct judgments, respectively), we also report Gwet’s AC1 alongside accuracy as a more conservative measure of LLM-as-judge performance.
\begin{table}[ht]
\centering
\begin{tabular}{lccc}
\hline
\textbf{Model} & \textbf{Accuracy} & \textbf{Gwet's AC1} \\
\hline
\textbf{LLaMA 3.1 8B Instruct}  & 0.92 & 0.86 \\
\textbf{Mistral 7B Instruct v0.3} & 0.88 & 0.76 \\
\textbf{Qwen3-32B} & 0.87 & 0.82 \\
\textbf{Phi-4} & 0.91 & 0.85 \\
\hline
\end{tabular}
\caption{The performance of the LLM-as-judge across different models' responses.}
\label{table:llm-as-judge-performance}
\end{table}

\end{document}